\documentclass[aip,jmp,amsmath,amssymb,reprint]{revtex4-2}

\usepackage[utf8]{inputenc}
\usepackage{amsmath,amssymb} %
\usepackage{graphicx} %
\usepackage{xcolor}
\usepackage{float}
\usepackage[left=1in, right=1in]{geometry}
\usepackage[utf8]{inputenc}
\usepackage{float}
\usepackage{textcomp,gensymb}
\usepackage[left=1in, right=1in]{geometry}
\usepackage{multirow}
\usepackage{mathtools}
\usepackage{ mathrsfs }

\usepackage{enumitem}
\newlist{todolist}{itemize}{2}
\setlist[todolist]{label=$\square$}
\usepackage{pifont}
\usepackage{lipsum}
\usepackage{amsfonts}
\usepackage{epstopdf}
\usepackage{algorithmic}
\usepackage{wrapfig}

\newcommand{\pers}{\textrm{pers}}

\renewcommand{\epsilon}{\varepsilon}
\renewcommand{\phi}{\varphi}

\newcommand{\boldA}{\mathbf{A}}

\newcommand{\R}{\mathbb{R}}

\graphicspath{{./Figures/}{./Figures/}}

\usepackage{graphicx}%
\usepackage{dcolumn}%
\usepackage{bm}%

\begin{document}

\preprint{AIP/123-QED}

\title{Persistent Homology of Coarse Grained State Space Networks}%

\author{Audun D.~Myers}
 \email{myersau3@msu.edu}
 \homepage{audunmyers.com}

\author{Max M.~Chumley}
 \email{chumleym@msu.edu}
  \homepage{maxchumley.com}
 
\author{Firas A.~Khasawneh}%
  \email{khasawn3@egr.msu.edu}
 \homepage{firaskhasawneh.com}
 \thanks{Corresponding Author}
 \affiliation{Department of Mechanical Engineering, Michigan State University, East Lansing, MI.}%

\author{Elizabeth Munch}%
 \email{muncheli@msu.edu}
 \homepage{elizabethmunch.com}
 \affiliation{Dept.~of Computation Mathematics Science and Engineering; and Dept.~of Mathematics, Michigan State University, East Lansing, MI}%

\date{\today}%

\begin{abstract}
This work is dedicated to the topological analysis of complex transitional networks for dynamic state detection. Transitional networks are formed from time series data and they leverage graph theory tools to reveal information about the underlying dynamic system. However, traditional tools can fail to summarize the complex topology present in such graphs. In this work, we leverage persistent homology from topological data analysis to study the structure of these networks. 
We contrast dynamic state detection from time series using CGSSN and TDA to two state of the art approaches:  Ordinal Partition Networks (OPNs) combined with TDA, and the standard application of persistent homology to the time-delay embedding of the signal. 
We show that the CGSSN captures rich information about the dynamic state of the underlying dynamical system as evidenced by a significant improvement in dynamic state detection and noise robustness in comparison to OPNs. 
We also show that because the computational time of CGSSN is not linearly dependent on the signal's length, it is more computationally efficient than applying TDA to the time-delay embedding of the time series.
\end{abstract}
\maketitle

\section{Introduction} \label{sec:intro}

Signal processing has been successfully and widely utilized to extract meaningful information from time series of dynamical systems including 
dynamic state detection~\cite{Gottwald2009,Gottwald2016,Melosik2016,Rosenstein1993,Froeschle1997,Sprott2004}, 
structural health monitoring for damage detection~\cite{Aval2019,Nejad2020,Guo2011,Sohn2001,Cao2017}, 
and biological health monitoring~\cite{Lee2014,Roberts2001,Liu2017,Frank2006,Khan2013,Millette2011,Omidvar2021}. 
A promising direction for signal processing is through studying the shape of signals. 
This is done by implementing tools from Topological Data Analysis (TDA)\cite{Dey2021,Munch2017} to study the shape of the attractor of the underlying dynamical system. 
This field of signal processing is known as Topological Signal Processing (TSP)~\cite{Robinson2014}, which has had many successful applications, including 
biological signal processing~\cite{Topaz2015,McGuirl2020}, 
dynamic state detection~\cite{Myers2020b,Myers2019}, 
manufacturing~\cite{Yesilli2020,Yesilli2020a,Yesilli2021,Yesilli2022,Yesilli2022a}, 
financial data analysis~\cite{Gidea2018,Gidea2020,Gidea2017a,Yen2021}, 
video processing~\cite{Hu2019,Tralie2019}, 
bifurcation detection~\cite{Tymochko2020}, 
and weather analysis~\cite{Tymochko2019,Muszynski2019}.

The standard pipeline for TSP constructs a filtration of simplicial complexes (called the Vietoris-Rips complex) based on point cloud data generated from the State Space Reconstruction (SSR) of an input time series~\cite{Chung2021,Emrani2014,Maletic2016,Yesilli2022}. 
Given a uniformly sampled signal $x = [x_1, x_2, \ldots, x_L]$, the SSR (also called the delay embedding) 
consists of $n$-dimensional delayed vectors 
\begin{equation}
    \mathbf{X} = \{v_i = [x_i, x_{i+\tau}, x_{i+2\tau}, \ldots, x_{i+\tau (n-1)}] 
    \mid  i \in \{1, \cdots, L - \tau (n-1)\} \}.
    \label{eq:SSR}
\end{equation}
A simplicial complex is formed by including simplices for all collections of points which are within distance $r$ of each other.
We can measure the shape of the simplicial complex by forming simplicial complexes at increasing values of $r$, and tracking the changing homology through a linear mapping. This allows for quantifying when specific topologies form and disappear throughout the filtration giving a sense of shape.
The persistence diagram encodes this information for various dimensions, e.g., connected components (dimension zero), loops (dimension one), voids (dimension two). For example, one can examine the one dimensional homology to track loop structures in the SSR that are related to the periodicity of the signal. 
A problem with this pipeline is its computational demand having complexity $O(N^3)$, where $N = \binom{n}{d+1}$ is the size of the simplicial complex with $n$ as the number of points in the simplical complex and $d$ as the maximum dimension of the used homology.  
For long signals, this makes this standard pipeline computationally infeasible. 
A common solution is to subsample the point cloud, but it can be challenging to select an appropriate subsampling rate that preserves the topology of interest. 

An alternative, promising direction for signal processing is analyzing time series via representations as complex networks~\cite{Small2009,Yang2008,Gao2009a}.
Network representations of time series generally fall within three categories: proximity networks, visibility graphs, and transitional networks.
Proximity networks are formed from closeness conditions in the reconstructed state space. 
Examples include the $k$-Nearest Neighbors ($k$-NN)~\cite{Khor2016} and recurrence networks~\cite{Donner2010}, where the recurrence network is the network underlying the Vietoris-Rips complex of the point cloud data.
For proximity networks, the graph representation includes all points in the state space reconstruction as part of the vertex set and does not reduce the computational complexity. 
Additionally, these networks require choosing a proximity parameter dependent on the signal, where careful consideration is needed in selecting the number of neighbors $k$ or proximity distance $\epsilon$ to generate a graph that captures the correct topology.
The visibility graphs \cite{Lacasa2008} are formed by adding vertices for each data point and adding connecting edges if a visibility line can be drawn between the two vertices which do not pass below any other data point between the two.
As our focus in this work is on building upon the strong theory developed for the SSR embedding, we will not utilize the visibility graph constructions at this stage. 
Instead, in this work we focus on transitional networks.

\begin{figure}%
    \centering
    \includegraphics[width = 0.48\textwidth]{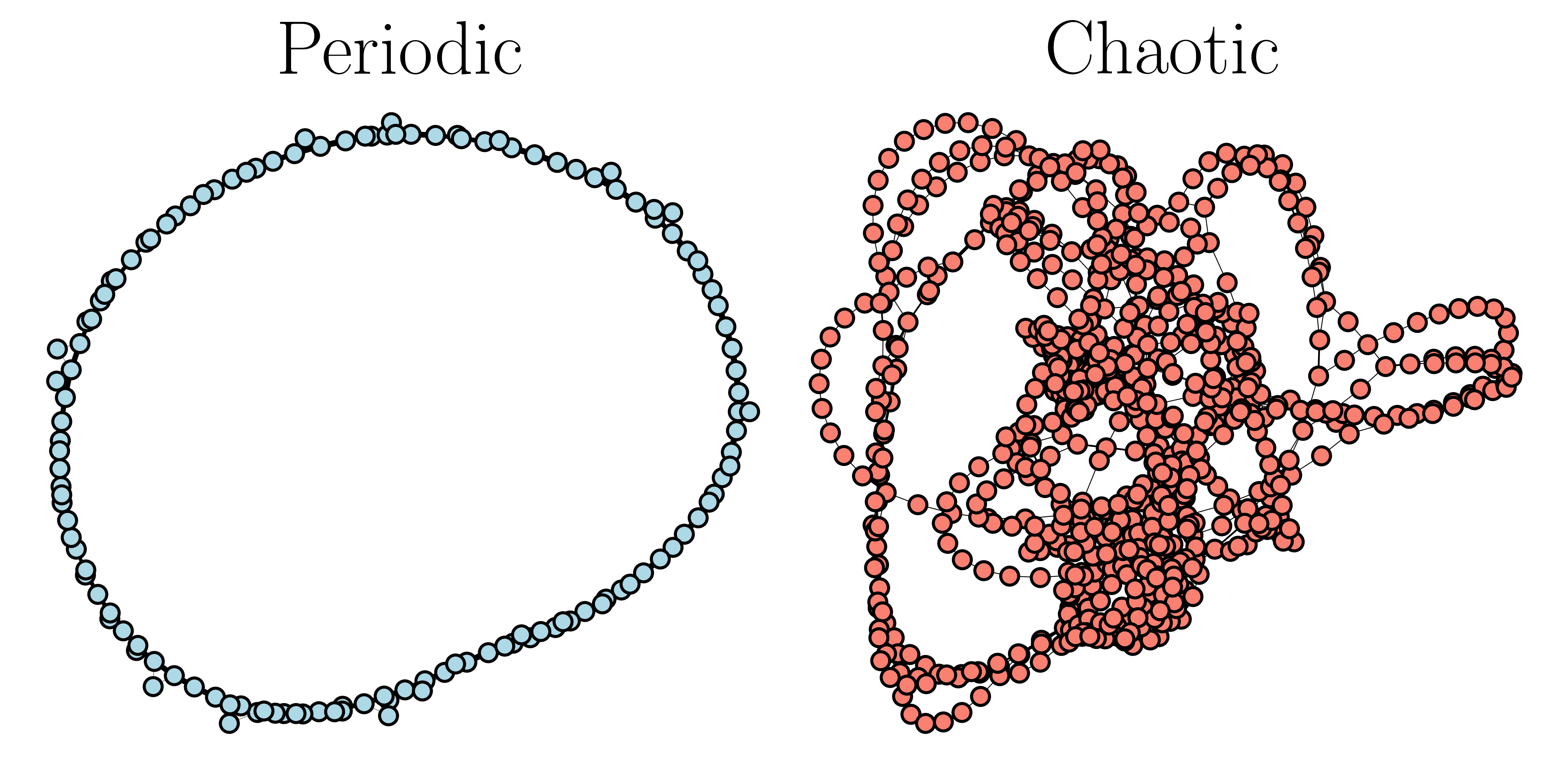}
    \caption{Example periodic and chaotic CGSSNs generated from the $x(t)$ solution to the R\"{o}ssler system.}
    \label{fig:chaotic_vs_periodic_networks}
\end{figure}

Transitional networks partition a time series $x$ such that it has a vertex set of states $\{s_i\}$ for each visited state and an edge for temporal transitions between states.
The resulting transitional network constitutes a finite state space $\mathcal{A}$ as the alphabet of possible states.
One interpretation of a topological system on a finite state space is as a finite graph where the edges describe the action of a function $\phi$, i.e., if there is a directed edge from vertex $a$ to vertex $b$, then $\phi(a)=b$.
Therefore, the transitional networks we obtain from a time series are topological systems, and they yield themselves to further analysis within the framework of topological dynamics. 
The two most common transitional networks for time series analysis are the Ordinal Partition Network (OPN) \cite{McCullough2015} and the Coarse Grained State Space Network (CGSSN) \cite{Wang2016,Weng2017,Campanharo2011,nicolis2005}. 
In Fig.~\ref{fig:chaotic_vs_periodic_networks} we demonstrate the rich topological structure of the CGSSN for periodic and chaotic dynamics from the Rossler system. This example shows the periodic dynamics corresponding to an approximate cycle graph while the network of the chaotic signal is highly intertwined.

To date, the majority of evaluation of these complex network representations is through standard graph theory tools~\cite{McCullough2015,Silva2022,Small2013,Gao2009a}, but the results can only provide local structural measurements based on the node degree distribution or shortest path measurements. 
In our previous work~\cite{Myers2019}, we studied the global shape of these networks using persistent homology for dynamic state detection using the ordinal partition network. 
However, we only used the shortest unweighted path to define distances between nodes, which discarded edge weight and direction information. 
In our recent work~\cite{Myers2022}
we investigate the use of weighted edge information based on the number of edge transitions. We found that this improved dynamic state detection performance.

However, we show here that there is an issue with the OPN; namely, amplitude information is discarded because the ordinal partition network is built from permutations. 
Permutations can be thought of as partitioning the state space via intersections of hyperplanes of the form $x_i \leq x_j$. 
As such, the resulting OPN can have reduced dynamic state detection performance and extreme sensitivity to additive noise for some signals. 
This can be partially explained by noting that proximity of the trajectory to the hyperdiagonal can cause failures in network construction, particularly when there is noise in the signal (details of this issue are provided in Section~\ref{ssec:noise_sensitivity}). 
Further, due to the hyperdiagonal intersection issue, we cannot guarantee the stability of the persistence diagram for all signals. 
Therefore, we turn our attention to the CGSSN to bypass the limitations in OPN. 

We investigate the applicability of the CGSSN for enhanced noise robustness and dynamic state detection compared to the OPN. 
The results presented are based on analyzing the  complex networks using persistent homology and tools from information theory and machine learning.
Our results show an improvement in dynamic state detection performance with 100\% separation between periodic from chaotic dynamics for noise-free signals using a nonlinear support vector machine compared to at most 95\% for the OPN. 
Additionally, we show an improved noise robustness with the CGSSN functioning down to a signal-to-noise ratio of 22 dB compared to 29 dB for the OPN. 

\subsubsection*{Organization}
In Section~\ref{sec:background} we overview the necessary background information. 
We begin with an introduction to the two transitional networks we study---OPN and CGSSN---and an overview of how they are related to state space reconstruction. 
Next, we introduce four standard methods for measuring the distance between nodes in a weighted graph. 
We subsequently describe persistent homology and how it is applied to study the shape of the weighted complex networks.
In Section~\ref{sec:method}, we demonstrate how to apply our pipeline for studying the shape of complex transitional networks for a simple periodic example.
In Section~\ref{sec:results}, we show results for studying the persistent homology of both the OPN and CGSSN. 
We begin with results for dynamic state detection for the Lorenz system with a periodic and chaotic response. 
We then apply the method to 23 continuous dynamical systems, and utilize machine learning to quantify the dynamic state detection performance over a broad range of signals. 
Lastly, we show results on the noise robustness of the CGSSN in comparison to the OPN.
In Section~\ref{sec:conclusion}, we provide conclusions future work on applying persistent homology to study the structure of transitional networks.

\section{Background} \label{sec:background}

\subsection{Transitional Complex Networks} \label{ssec:transitional_networks}
A graph $G = (V,E)$ is a collection of vertices $V$ and edges $E = (u, v) \subseteq V \times V$.
We assume all graphs are simple (no self-loops or hypergraphs) and undirected. 
Additional stored information comes as a weighted graph, $G = (V,E, \omega)$ where $\omega:E \to \R_{\geq 0}$ gives a non-negative weight for each edge in the graph.
Given an ordering of the vertices $V = \{v_1,\cdots, v_n\}$, a graph can be stored in an adjacency matrix $\boldA$ where the weighting information is stored by setting $\boldA_{ij} = w_{(v_i, v_j)}$ if $(v_i, v_j) \in E$ and 1 otherwise. 

Transitional networks  are graphs  formed from a chronologically ordered sequence of symbols or states derived from the time series data.
In our construction, these states are mapped from the measurement signal by first creating an SSR $\mathbf{X}$ from Eq.~\eqref{eq:SSR} and then assigning a symbolic representation for each vector $v_i \in \mathbf{X}$. 
To form a symbolic sequence from the time series data, we implement a function to map the SSR to symbol in the alphabet $\mathcal{A}$ of possible states as $f: v_i \rightarrow s_j$, where $s_j \in \mathcal{A}$ is a symbol from the alphabet. In this work, we consider the symbols from the alphabet as integers such that $s_i \in \mathbb{Z} \cap [1, N]$, where $N$ is the number of possible symbols. Applying this mapping over all embedding vectors we get a symbol sequence as $S = [s_1, s_2, \ldots, s_{L-\tau (n-1)}]$.
This work investigates two methods for mapping SSR vectors $v_i$ to symbols $s_j$. The first is the OPN which is defined in Section~\ref{sssec:OPN} and is based on permutations. The second method is the CGSSN defined in section~\ref{sssec:CGSSN} which uses an equal-sized hypercube tessellation. 

The symbol sequence $S$ forms a transitional network by considering a graph $G = (V, E)$, where the vertices $V$ are the collection of the used symbols, and the edges are added based on transitions between symbols in $S$.
We represent the graph using the adjacency matrix $\boldA$ data structure of size $N \times N$.
We add edges to the adjacency matrix $\boldA$ via the symbolic transitions with an edge between row $s_i$ and column $s_{i+1}$ for each $i$.
This is represented in the adjacency matrix structure by incrementing the value of $\boldA_{s_i, s_j}$ by one for each transition between $s_i$ and $s_{s+1}$, where $\boldA$ begins as a zero matrix.
\begin{figure}
    \centering
    \includegraphics[scale = 0.85]{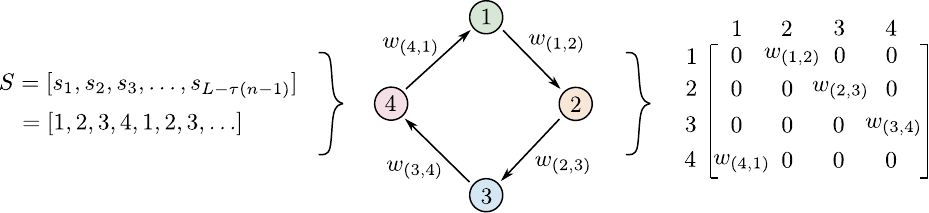}
    \caption{Example formation of a weighted transitional network as a graph (middle figure) and adjacency matrix (right figure) given a state sequence $S$ (left figure).}
    \label{fig:transition_network_formation_example}
\end{figure}
We set the total number of transitions between two nodes as the edge weight $w_{(s_i, s_j)}$. We ignore self-loops by setting the diagonal of $\boldA$ to zero.
To better illustrate the transitional network formation process, consider the simple cycle shown in Fig.~\ref{fig:transition_network_formation_example}. 
In this example, we take the state sequence $S$ on the left side of Fig.~\ref{fig:transition_network_formation_example} with symbols in the alphabet $\mathcal{A} = [1, 2, 3, 4]$ and create the network shown network in the middle of the figure. 
This network is represented as a directed and weighted adjacency matrix, as shown on the right side of Fig.~\ref{fig:transition_network_formation_example}. 
In this paper, we discard the directionality information and make $\boldA$ symmetric by adding its transpose, $\boldA + \boldA^T$.

\subsubsection{Ordinal Partition Network} \label{sssec:OPN}
\begin{figure}[htbp]
    \centering
    \includegraphics[width = 0.97\textwidth]{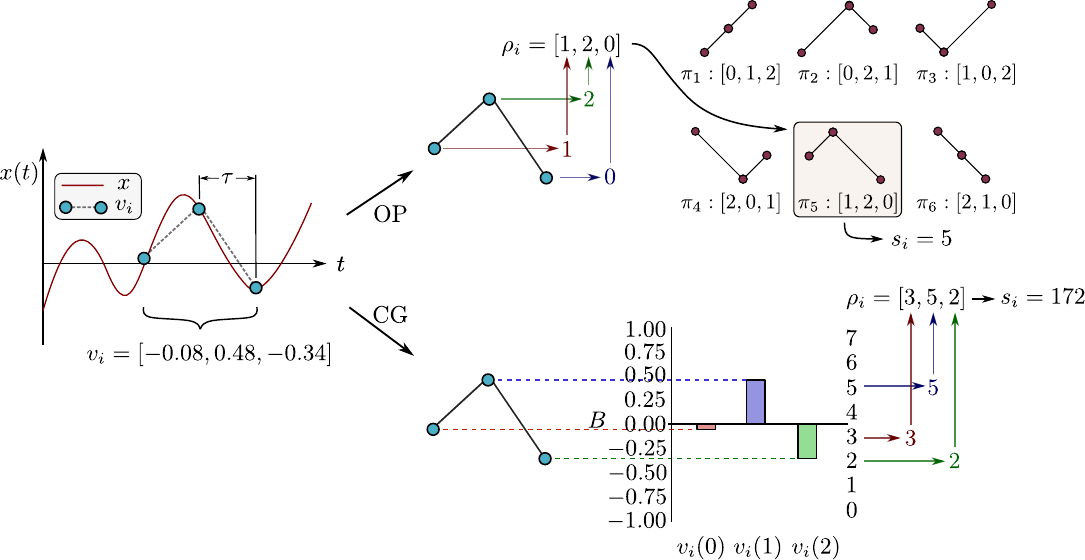}
    \caption{Example state assignment using the Ordinal Partition (OP) method (top) and Coarse Graining (CG) method (bottom). The state for the OP method is based on the assigned permutation number with $s_i = 5$ for the example. The state assignment for the CG method is based on the number of bins where $s_i = 1 + \sum_{j = 0}^{n-1} \rho_i(j) b^j$, $\rho_i$ is the digitization of vector $v_i$ based on binning into $b$ equal-sized bins spanning $[\min(x), \max(x)]$. For this example, $s_i = 3(8^0) + 5(8^1) + 2(8^2) + 1 = 172$ with $b=8$ bins.}
    \label{fig:state_assignment_OPN_and_CGSSN}
\end{figure}

To form an OPN, the SSR $\mathbf{X}$ must first be constructed requiring the choice of two parameters: the delay $\tau$ and dimension $n$. We select the delay $\tau$ using the method of multi-scale permutation entropy~\cite{Staniek2007,Myers2020a} and the dimension as $n=7$ as suggested for permutation entropy~\cite{Myers2020a,Myers2020a}.
For the OPN, the vector $v_i$ is assigned to a permutation $\pi$ based on its ordinal partition. 
For dimension $n$ there are $n!$ permutations (e.g., 6 possible permutations for dimension $n=3$ shown in 
Fig.~\ref{fig:state_assignment_OPN_and_CGSSN}) which can order arbitrarily $\pi_1,\cdots \pi_{n!}$. 
Then $v_i$ is assigned to a permutation $\pi_k$ following that $\pi_k$ satisfies $v_i(\pi_k(0)) \leq v_i(\pi_k(1)) \leq \ldots \leq v_i(\pi_k(n-1))$. 
An example of this for the vector $v_i = [-0.08, 0.48, -0.34]$ is shown on the top Ordinal Partition (OP) route of Fig.~\ref{fig:state_assignment_OPN_and_CGSSN} where $v_i$ is mapped to permutation $\pi_5$ and state $s_i = 5$.
\subsubsection{Coarse Grained State Space Network} \label{sssec:CGSSN}
The CGSSN begins by constructing the SSR, where we select the delay $\tau$ using the multi-scale permutation entropy method~\cite{Staniek2007,Myers2020a} and dimension $n$ using the false nearest neighbors~\cite{Krakovska2015} based on only needing a dimension great enough for periodic orbits to not self-intersect.
For the CGSSN, the vector $v_i \in \mathbf{X}$ is assigned to a state based on which partitioned region the vector $v_i$ lies within. 
We define the domain $\mathbb{D}$ of the SSR as the non-empty connected, open set that encloses all vectors of the SSR. 
Specifically, we use an $n$-dimensional hypercube domain bounded by the intervals $[\min(x), \max(x)]$ for each dimension. 
In this work we cover this domain using a tessellation of $N = b^n$ hypercubes with side length $(\max(x) - \min(x))/b$, where $b$ is the number of bins per dimension. 
We assign each $n$-dimensional hypercube in the tessellation a unique symbol by converting it to a decimal representation denoted as $s_i$. An introductory example formation of the entire CGSSN for a sinusoidal function is provided in Section~\ref{sec:method}. Some generalizations exist to the described method where instead of assigning symbols to the individual hypercubes, we could assign words of length $m$ which would allow for studying a sequence of coarse grained states of the system which reduces the information load in the process \cite{zou2019}. For the purpose of this paper, a symbolic representation was sufficient.

\subsection{Vertex similarity and dissimilarity measures} 
\label{ssec:distances_measures}

To study the structure of the complex network we define functions of the form $V \times V \to \R_{\geq 0}$ combining information about path lengths and weights from the graph in various ways.  
Some of these definitions are distances, but not all. 
Despite this, the framework can still be used to define a filtered simplicial complex in the spirit of the Vietoris Rips complex which will be required in the next section.  

The measures are encoded in a matrix $\mathbf{D}$, where $\mathbf{D}(a,b)$ is the similarity or dissimilarity between vertices $a$ and $b$. 
Note that $\mathbf{D}$ can optionally be normalized by dividing all entries by its maximum value to contain values between 0 and 1.
We investigated the use of four choices of measures: the unweighted shortest path distance, the shortest weighted path dissimilarity, the weighted shortest path distance, and the diffusion distance.

\subsubsection{Shortest Path Distances and Dissimilarities}
\label{sssec:shortestPath}
Commonly used in graph theory, the \textit{shortest path distance} is based on minimizing the cost of taking a path from node $a$ to $b$. 
This assumes a path $P = [n_0, n_1, \ldots, n_s]$ consisting of $s$ nodes where $a = n_0$ and $b=n_s$ exists, but we note that all graphs in this paper are connected by construction. 
The path $P$ can alternatively be represented as the sequence of connected edges between $a$ and $b$: $P = [e_{0,1}, e_{1,2}, \ldots, e_{s-1,s}]$. 
The shortest path is determined based on minimizing the path cost function
\begin{equation} \label{eq:shortest_path_cost}
C(P) = \sum_{e \in P} w(e).
\end{equation}
In the case of an weighted graph, we then define $D(a,b) = \min_P C(P)$. 
Note that in the case of an unweighted graph, we have all weights equal to 1 and thus the cost of a path is simply the number of edges included in it. 

We next define two variations on this idea, although they are not quite distances but are useful for the kinds of input graph data we study. 
In particular, the weights on edges are higher for those that are more highly traversed with the transitional networks. 
We thus want these paths to be considered more important than those only traversed a few times. 
To that end, we will focus on paths whose length using the reciprocal of the weights is as small as possible. 

The first variation, called the \textit{weighted shortest path} measure, is defined as follows. 
First, we find the path from $a$ to $b$ with the minimum total path weight in terms of the reciprocal weights. 
That is, $P$ such that 
\begin{equation} %
C'(P) = \sum_{e \in P} 1/w(e).
\end{equation}
is minimized.
We then define $D(a,b) = \sum_{e \in P} w(e)$. 
For this definition, $D$ encodes information about frequency of traversal of the edges. 

The second variation, called the \textit{shortest weighted path}, still uses the path $P$ for which $C'(P)$ is minimized.  
However, in this case, we define $D(a,b)$ to be the length of the path; i.e.~the number of edges in $P$. 
For this variant, we are essentially giving higher priority to well traveled paths, but using a measurement of this path related to the number of regions of state space are traversed.

\subsubsection{Diffusion Distance}
The final vertex similarity measure we use is the diffusion distance for graphs~\cite{Coifman2006}.
The diffusion distance leverages the transition probability distribution matrix $\mathbf{P}$ of the graph, where $\mathbf{P}(a,b)$ is the probability of transitioning to $b$ when at $a$ in a single step based on the random walk framework.
Specifically, given the weighted, undirected adjacency matrix $\mathbf{A}$ with no self-loops (i.e., zero diagonal), the transitional probability matrix is 
\begin{equation} \label{eq:prob_dist_mat}
\mathbf{P}(i,j) = \frac{\mathbf{A}{(i,j)}}{\sum_{k=1}^{|V|} \mathbf{A}{(i,k)}}.
\end{equation}
Equation~\eqref{eq:prob_dist_mat} can be extended to calculate the transition probabilities for non-adjacent neighbors by raising them to higher powers. For example, transitioning to vertex $b$ from vertex $a$ in $t$ random walk steps is $\mathbf{P}^t(a, b)$. A common modification of Eq.~\eqref{eq:prob_dist_mat} is to include a probability that a random walk can stay at the current vertex, which is commonly referred to as the lazy transition probability matrix. This is given by
\begin{equation}  \label{eq:lazy_prob_mat}
\widetilde{\mathbf{P}} = \frac{1}{2}\left[\mathbf{P}(a, b) + \mathbf{I}\right],
\end{equation}
where $\mathbf{I}$ is the identity matrix matching the size of $\mathbf{P}$.
The diffusion distance measures how similar two nodes are based on comparing their $t$-step random walk probability distributions. This is done by taking the degree-normalized $\ell_2$ norm of the probability distributions between nodes and is calculated as
\begin{equation}
	d_t(a,b) = \sqrt{ \sum_{c \in V}  \frac{1}{\mathbf{d}(c)} { \left[ \widetilde{\mathbf{P}}^t (a,c) - \widetilde{\mathbf{P}}^t (b,c) \right] }^2 }
\end{equation}
where $\mathbf{d}$ is the degree vector of the graph with $\mathbf{d}(i)$ as the degree of node $i$. Applying the diffusion distance to all node pairs results in the distance matrix $\mathbf{D}_t$.

\subsection{Persistent Homology of Complex Networks} \label{ssec:persistent_homology}

\begin{figure}
    \centering
    \begin{minipage}[c]{0.15\textwidth}
        \centering
        \includegraphics[width=\linewidth]{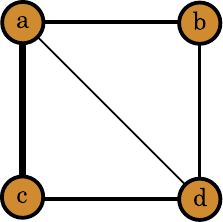} 
        \label{fig:G_toy_example}
    \end{minipage}
    \hspace{7mm}
    \begin{minipage}[c]{0.16\textwidth}
        \centering
        \includegraphics[width=\linewidth]{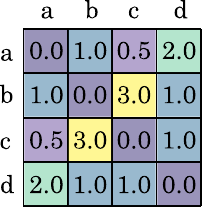} 
         {$\mathbf{D}$}
        \label{fig:D_toy_example}
    \end{minipage}
    \hspace{7mm}
    \begin{minipage}[c]{0.30\textwidth}
        \centering
        \includegraphics[width=\linewidth]{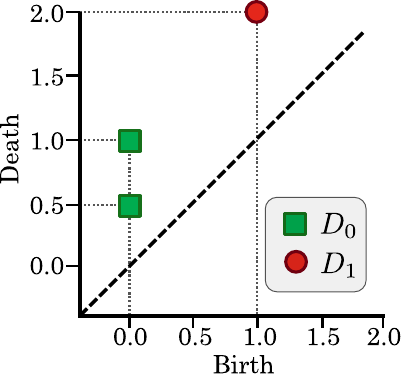} 
        \label{fig:PD_toy_example}
    \end{minipage}
    \vspace{2mm}
    
    \begin{minipage}[t]{0.15\textwidth}
        \centering
        \includegraphics[width=\linewidth]{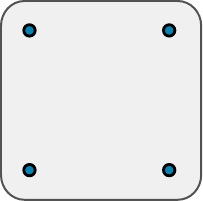} 
         {$K_{0.0}$}
        \label{fig:K0_toy_example}
    \end{minipage}
    \hspace{3mm}
    \begin{minipage}[t]{0.15\textwidth}
        \centering
        \includegraphics[width=\linewidth]{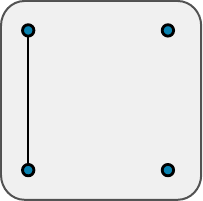} 
         {$K_{0.5}$}
        \label{fig:K1_toy_example}
    \end{minipage}
    \hspace{3mm}
    \begin{minipage}[t]{0.15\textwidth}
        \centering
        \includegraphics[width=\linewidth]{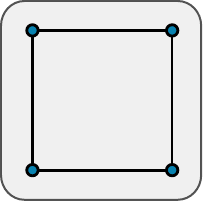} 
         {$K_{1.0}$}
        \label{fig:K2_toy_example}
    \end{minipage}
    \hspace{3mm}
    \begin{minipage}[t]{0.15\textwidth}
        \centering
        \includegraphics[width=\linewidth]{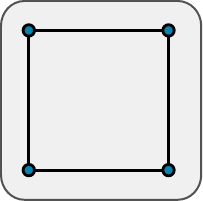} 
         {$K_{1.5}$}
        \label{fig:K3_toy_example}
    \end{minipage}
    \hspace{3mm}
    \begin{minipage}[t]{0.15\textwidth}
        \centering
        \includegraphics[width=\linewidth]{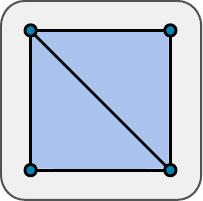} 
         {$K_{2.0}$}
        \label{fig:K4_toy_example}
    \end{minipage}
    \caption{Example demonstrating persistent homology of a graph  using the  matrix $\mathbf{D}$ with resulting persistence diagram shown top right. 
    The filtration of simplicial complexes are shown in the bottom row.}
    \label{fig:PH_toy_example}
\end{figure}

A simplicial complex is a generalization of a graph to higher dimensions,  which are collections of simplices at various dimensions (e.g., points are zero-dimensional, edges are one-dimensional, and faces are two-dimensional simplices). 
These simplices are subsets of a vertex set $\sigma \subset V$, and we require for the complex that if $\sigma \in K$ and $\tau \subseteq \sigma$, then $\tau$ is also in $K$. 
Using a distance matrix to describe similarity between nodes, or indeed any function of the form $d: V \times V \to \R$ where $d(v,v) = 0$ although we still call this a distance matrix for simplicity, we can construct simplicial complex representations from graphs at a distance level $r$. 
This idea is related to the Vietoris Rips complex, where we build a simplicial complex $K_r$ for any fixed parameter $r \geq 0$ by including all simplices with pairwise relationships at most $r$; i.e.~$K_r = \{ \sigma \subseteq V \mid d(u,v) \leq r \text{ for all } u,v \in \sigma\}$.
Zero-dimensional simplices, the vertices of the complex,  are all added at $r=0$.
An edge $uv$, which is a 1-dimensional simplex, is present in $K_r$ for any $r$ value above $d(u,v)$. 
Higher dimensional simplices such as triangles are included when all subedges are present; equivalently this means a simplex is added for every clique in the complex. 
For example, consider Fig.~\ref{fig:PH_toy_example} which shows a graph with four nodes, and the associated distance matrix $\mathbf{D}$. 
For each $r \in [0.0, 0.5, 1.0, 1.5, 2.0]$ the associated simplicial complex is shown as $K_r$ in  the bottom row. 

We can use homology~\cite{Hatcher2002,Munkres1993} to measure the shape of any such simplicial complex $K$ which is denoted $H_d(K)$.
This mathematical object is a vector space, where elements are representative of $d$-dimensional features (i.e., connected components (zero-dimensional structure), loops (one-dimensional structure), voids (two-dimensional structure), and higher dimensional analogues) in $K$.
In this work we will only utilize the 0-dimensional and 1-dimensional features to measure the connected components and holes in the simplicial complex.
For example, consider the simplicial complex $K_r$ at $r=1.0$ in Fig.~\ref{fig:PH_toy_example}, which has one $H_0$ classes with a single connected component and one $H_1$ class with a single loop or hole in the simplicial complex.

An issue with just using homology to measure the shape of a simplicial complex to understand the shape of a graph is that the correct distance value $r$ needs to be selected. 
Additionally, it does not provide any information on the geometry or size of the underlying graph. 
To alleviate these issue we use persistent homology~\cite{Zomorodian2004}, which studies the \textit{changing} homology of a sequence of simplicial complexes.
We will again use Fig.~\ref{fig:PH_toy_example} as an example for demonstrating how the persistent homology is calculated.
To calculate the persistent homology we begin with a collection of nested simplicial complexes
\begin{equation*}
  K_{r_1} \subseteq K_{r_2} \subseteq \cdots \subseteq K_{r_N}.
\end{equation*}
The bottom row of Fig.~\ref{fig:PH_toy_example} shows an example of this filtration over the distance parameter $r$ with $K_{r=1.0} \subseteq K_{r=0.5} \subseteq \cdots \subseteq K_{r=2.0}$.
We then calculate the homology of each simplicial complex and create linear maps between each homology class for each dimension $d$ as
\begin{equation*}
  H_d(K_{r_1}) \to H_d(K_{r_1}) \to \cdots \to H_d(K_{r_N}).
\end{equation*}
By studying the formation and disappearance of homology classes we can understand the shape of the underlying graph. Specifically, class $[\alpha] \in H_d(K_{r_i})$ is said to be born at $r_i$ if it is not in the image of the map $H_d(K_{r_{i-1}}) \to H_d(K_{r_i})$.
The same class dies at $r_j$ if $[\alpha] \neq 0$ in $H_d(K_{r_{j-1}})$ but $[\alpha] = 0$ in $H_d(K_{r_{j}})$.
In the case of 0-dimensional persistence, this feature is encoding the appearance of a new connected component at $K_{r_i}$ that was not there previously, and which merges with an older component entering $K_{r_j}$.
For 1-dimensional homology, this is the formation (birth) and disappearance (death) of a loop structure.
We store this information in what is known as the persistence diagram using the persistence pair $x_i = (b_i, d_i) \in D_d$, where $D_d$ is the persistence diagram of dimension $d$ with a homology class of dimension $d$ being born at filtration value $b_i$ and dying at $d_i$. We also define the lifetime or persistence of a persistence pair as $\ell_i = {\rm pers}(x_i) = d_i - b_i$. The set of lifetimes for dimension $d$ is defined as $L_d$.
For a more detailed roadmap for the calculation of persistent homology we direct the reader to the work of Otter et~al~\cite{Otter2017}.

Returning to our example, the persistence diagram is shown in Fig.~\ref{fig:PH_toy_example} for both $D_0$ and $D_1$. For $D_0$ all four persistence pairs were born at $r=0.0$ with one dying at $r=0.5$ and two dying at $r=1.0$. 
The fourth persistence pair in $D_0$, not drawn, is an infinite-class dying at $\infty$ since there is a single component for $r\geq 1.0$. 
In this work we do not utilize infinite-class persistence pairs and will not include them in the persistence diagrams. 
For $D_1$ there is a single persistence pair born at $r=1.0$ with the formation of the loop in $K_1$ and filling in at $K_2$.

\section{Method} \label{sec:method}

\begin{figure}
    \centering
    \includegraphics[width = 0.99\textwidth]{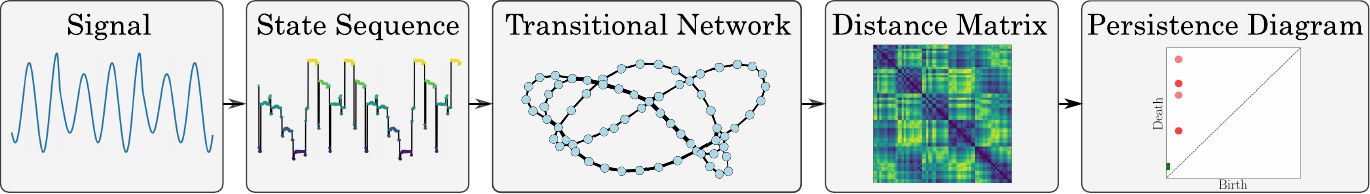}
    \caption{Pipeline for studying transitional networks using persistent homology. From left to right, we begin with a signal or time series and represent it as a state sequence which is summarized using a transitional network as described in Section~\ref{ssec:transitional_networks}. A distance between nodes is then used to create a distance matrix (see Section~\ref{ssec:distances_measures} for graph distances) which can be directly analyzed using persistent homology shown in Section~\ref{ssec:persistent_homology}.}
    \label{fig:method_pipeline}
\end{figure}
This section describes the method for studying complex transitional networks using persistent homology. The pipeline for doing this is outlined in Fig.~\ref{fig:method_pipeline}. We begin with a signal or time series and represent it as a state sequence described in Section~\ref{ssec:transitional_networks}. 
The state sequence can be summarized using a weighted transitional network as described in Sec.~\ref{ssec:transitional_networks}. 
A distance between nodes (see Section~\ref{ssec:distances_measures}) is then used to create a distance matrix which can be directly analyzed using persistent homology as described in Section~\ref{ssec:persistent_homology}.

\begin{figure}
    \centering
    \begin{minipage}[t]{0.59\textwidth}
    \centering
        \includegraphics[width = \linewidth]{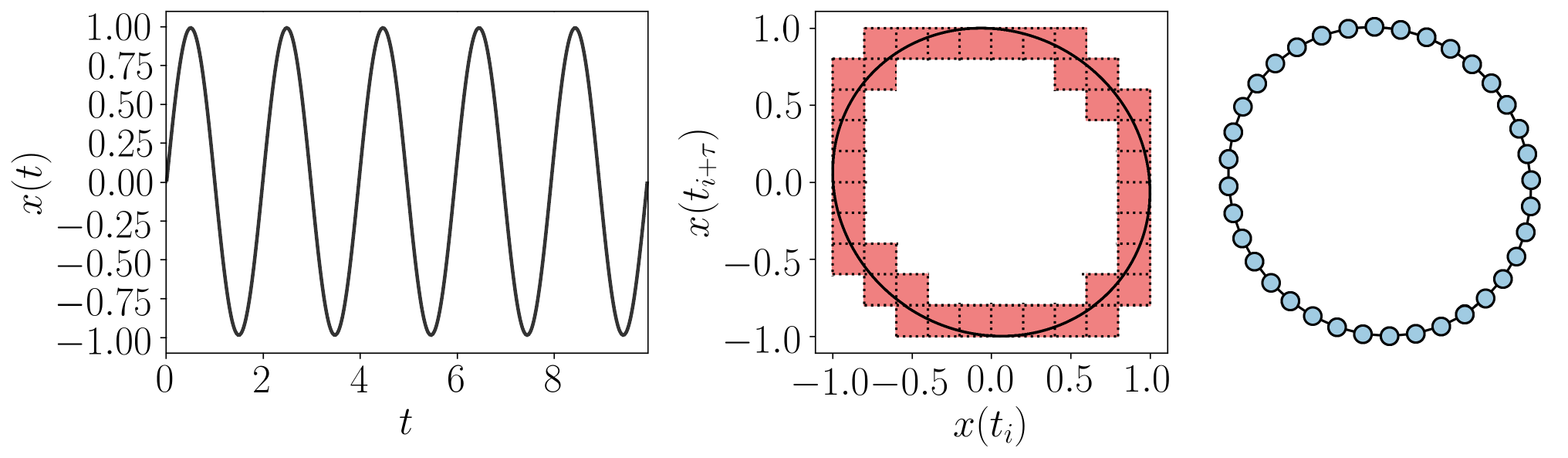}
        {\small{(a)}}
    \end{minipage}
    \begin{minipage}[t]{0.4\textwidth}
    \centering
        \includegraphics[width = \linewidth]{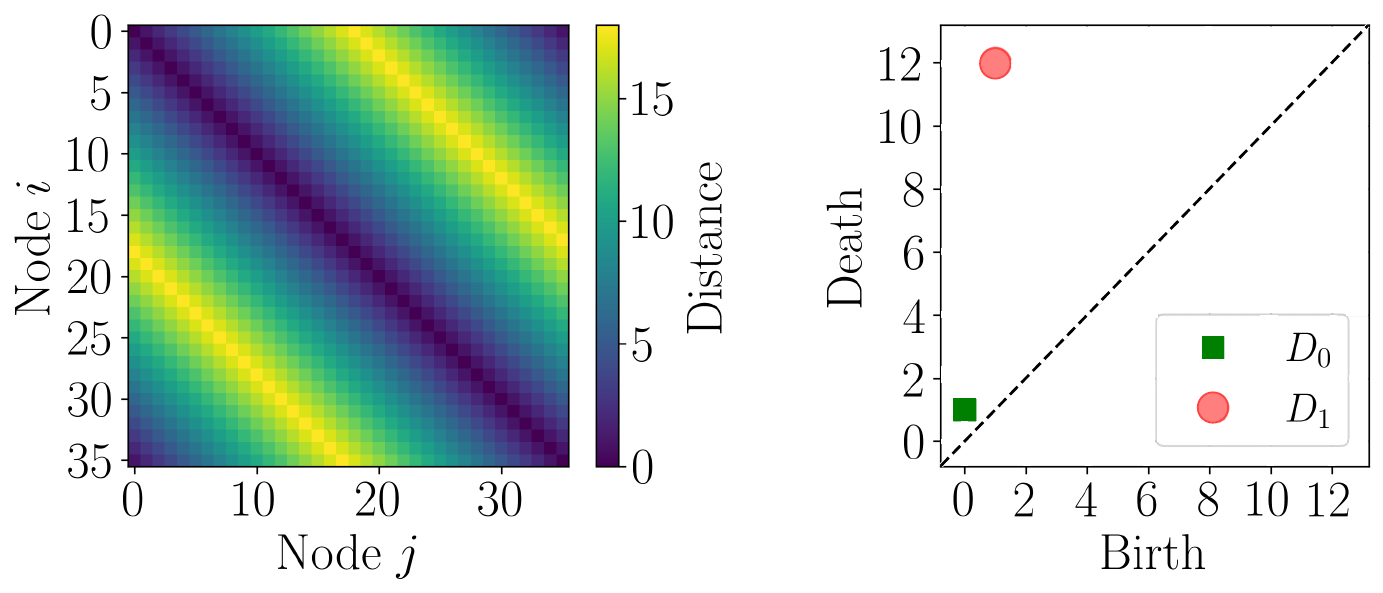}
        {\small{(b)}}
    \end{minipage}
    \caption{Example demonstrating CGSSN formation procedure ($b=10$) with the signal $x(t)= \sin(t)$ embedded into $\mathbb{R}^2$ space using an SSR and analysis using persistent homology with the unweighted shortest path distance. (a) Formation of the CGSSN from a time series signal and its delayed signal, (b) The distance matrix and associated persistence diagram using the unweighted shortest path distance.}
    \label{fig:example_periodic_2D_CGSSN_all}
\end{figure}

To further describe the method we develop here, we use a simple periodic signal example shown in Fig.~\ref{fig:example_periodic_2D_CGSSN_all}. 
The signal is defined as $x(t) = \sin(\pi t)$ sampled at a uniform rate of $f_s = 50$ Hz. 
The SSR was constructed using $n=2$ and $\tau = 26$. 
For this example, we create the CGSSN by partitioning the SSR domain into 100 rectangular regions as states, each with a unique symbol. 
The states visited through the SSR trajectory are highlighted in red. 
The temporal tracking of the states used creates the state sequence, which is then represented as the cycle graph. %
This example demonstrates how the periodic nature of the signal is captured by the cycle structure of the corresponding CGSSN.

We define a distance between nodes using the unweighted shortest path distance for this example due to its simplicity. 
The corresponding distance matrix and resulting persistence diagram are shown. %
The resulting persistence diagram shows that the periodic structure of the underlying time series and corresponding CGSSN is captured by the single point in the persistence diagram $D_1$ at coordinate $(1,12)$ with the loop structure being born at a filtration distance of 1 and filling in 12. 

\section{Results} \label{sec:results}
This section shows that the CGSSN outperforms the previously used OPN for both noise robustness and dynamic state detection performance. 
We first begin in Section~\ref{ssec:dynamic_state_detection_results_rossler} where we provide a simple example highlighting improved dynamic state detection performance of the CGSSN over the OPN for a periodic and chaotic Rossler system simulation. 
We show these results using the persistent entropy summary statistic. 
The second result in Section~\ref{ssec:dynamic_state_detection_23_systems} quantifies the dynamic state detection, of the OPN and CGSSN using lower dimensional embedding on 23 continuous dynamical systems with periodic and chaotic simulations. 
Lastly, in Section~\ref{ssec:noise_sensitivity}, we empirically investigate the noise robustness of the CGSSN compared to the OPN.

\subsection{Dynamic State Detection for Rossler System} \label{ssec:dynamic_state_detection_results_rossler}
\begin{figure}
    \centering
    \begin{minipage}[t]{0.35\textwidth}
        \centering
        \includegraphics[width=\linewidth]{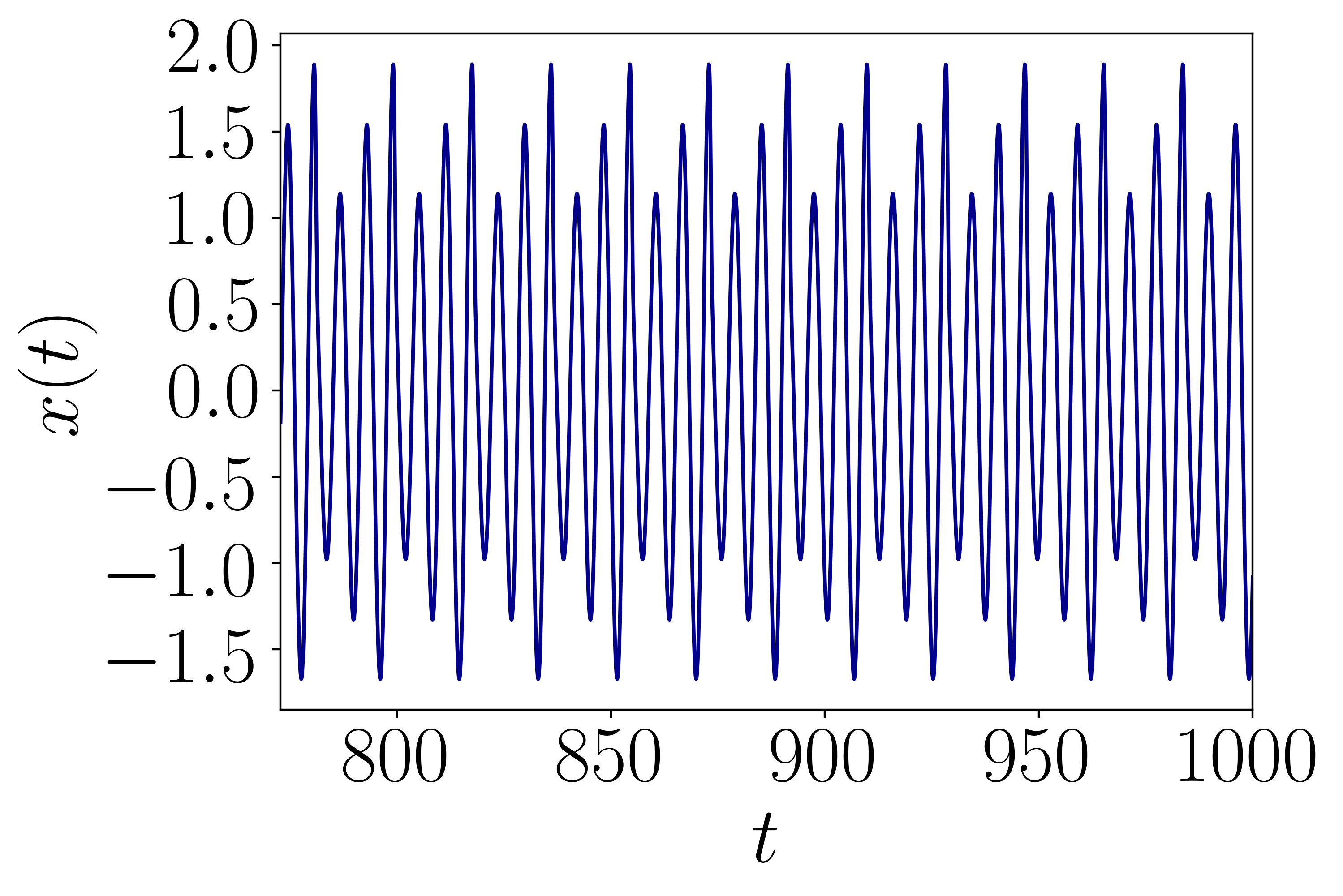} 
        {\small{(a)}}
    \end{minipage}
    \begin{minipage}[t]{0.28\textwidth}
        \centering
        \includegraphics[width=\linewidth]{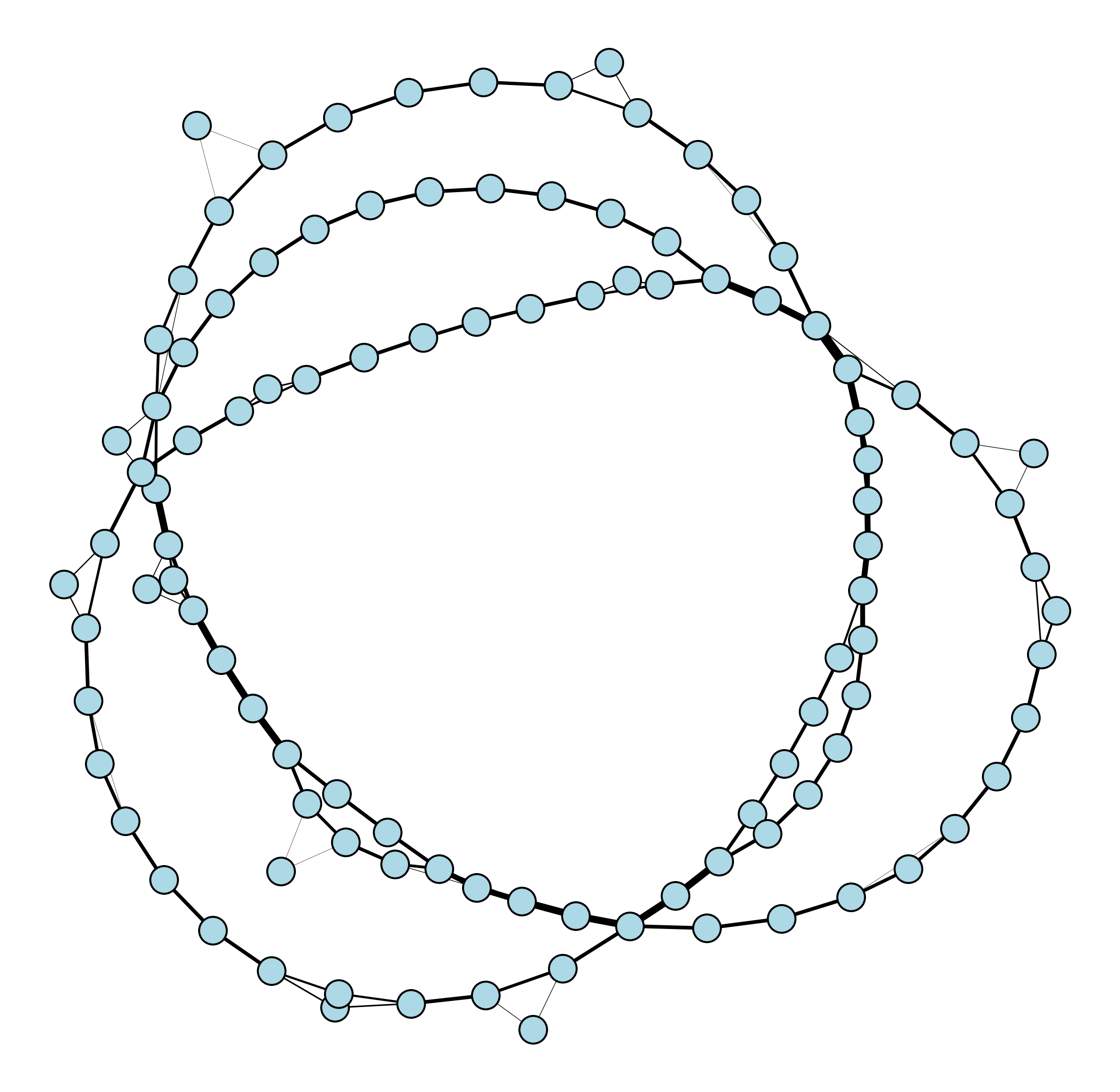} 
        {(b)}
    \end{minipage}
    \begin{minipage}[t]{0.28\textwidth}
        \centering
        \includegraphics[width=\linewidth]{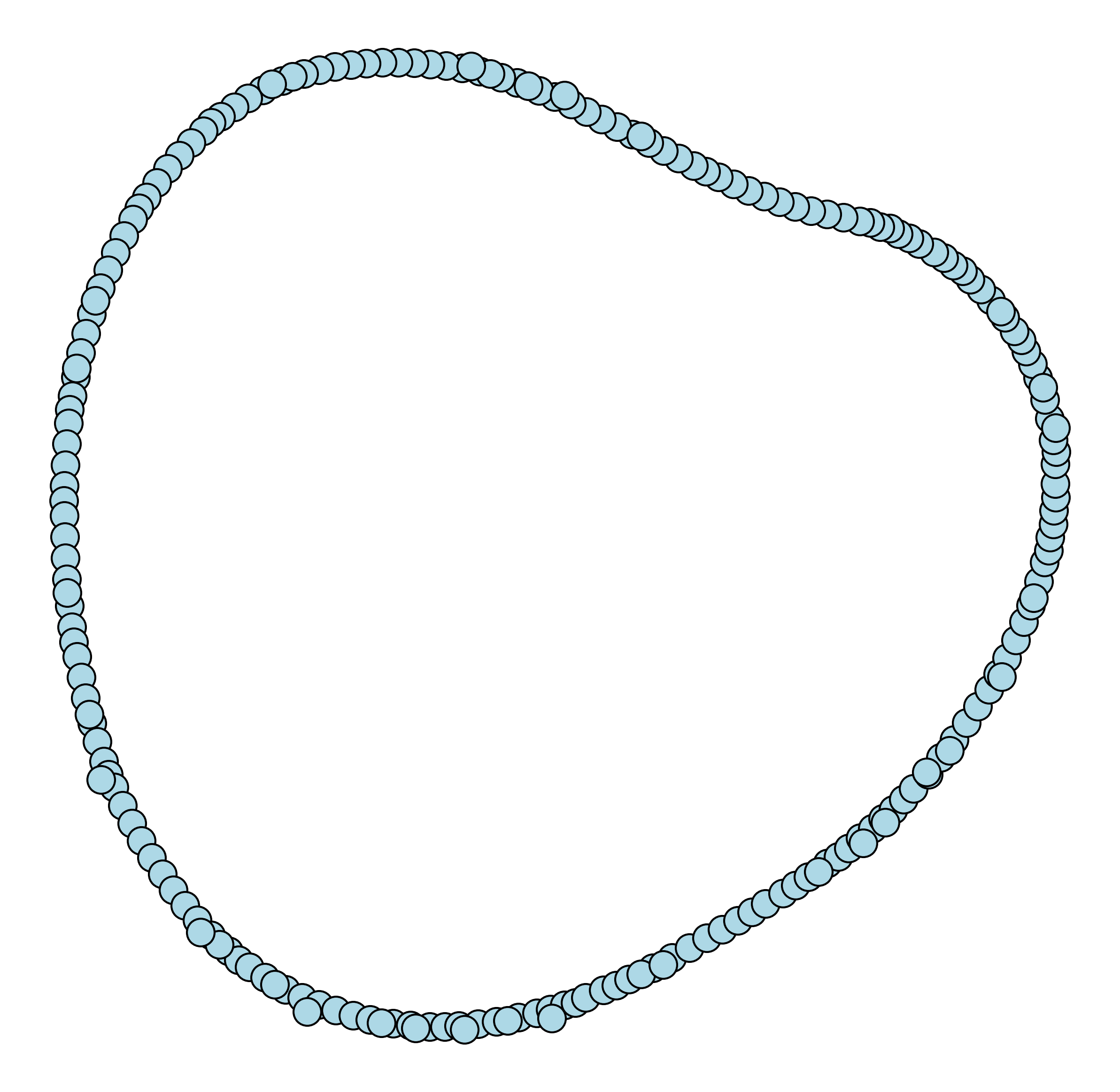} 
        {(c)}
    \end{minipage}

    \begin{minipage}[t]{0.35\textwidth}
        \centering
        \includegraphics[width=\linewidth]{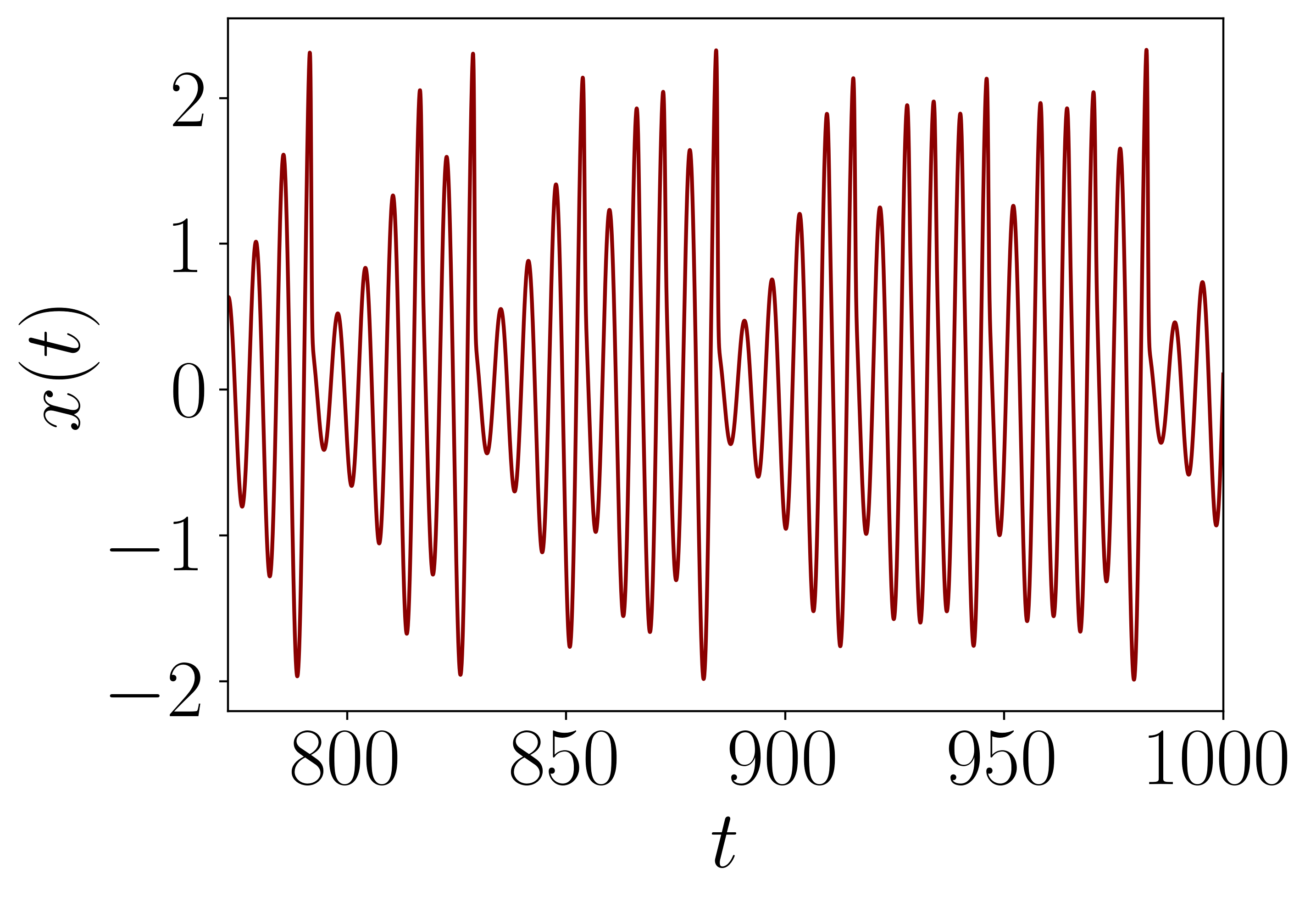} 
        {(d)}
    \end{minipage}
    \begin{minipage}[t]{0.28\textwidth}
        \centering
        \includegraphics[width=\linewidth]{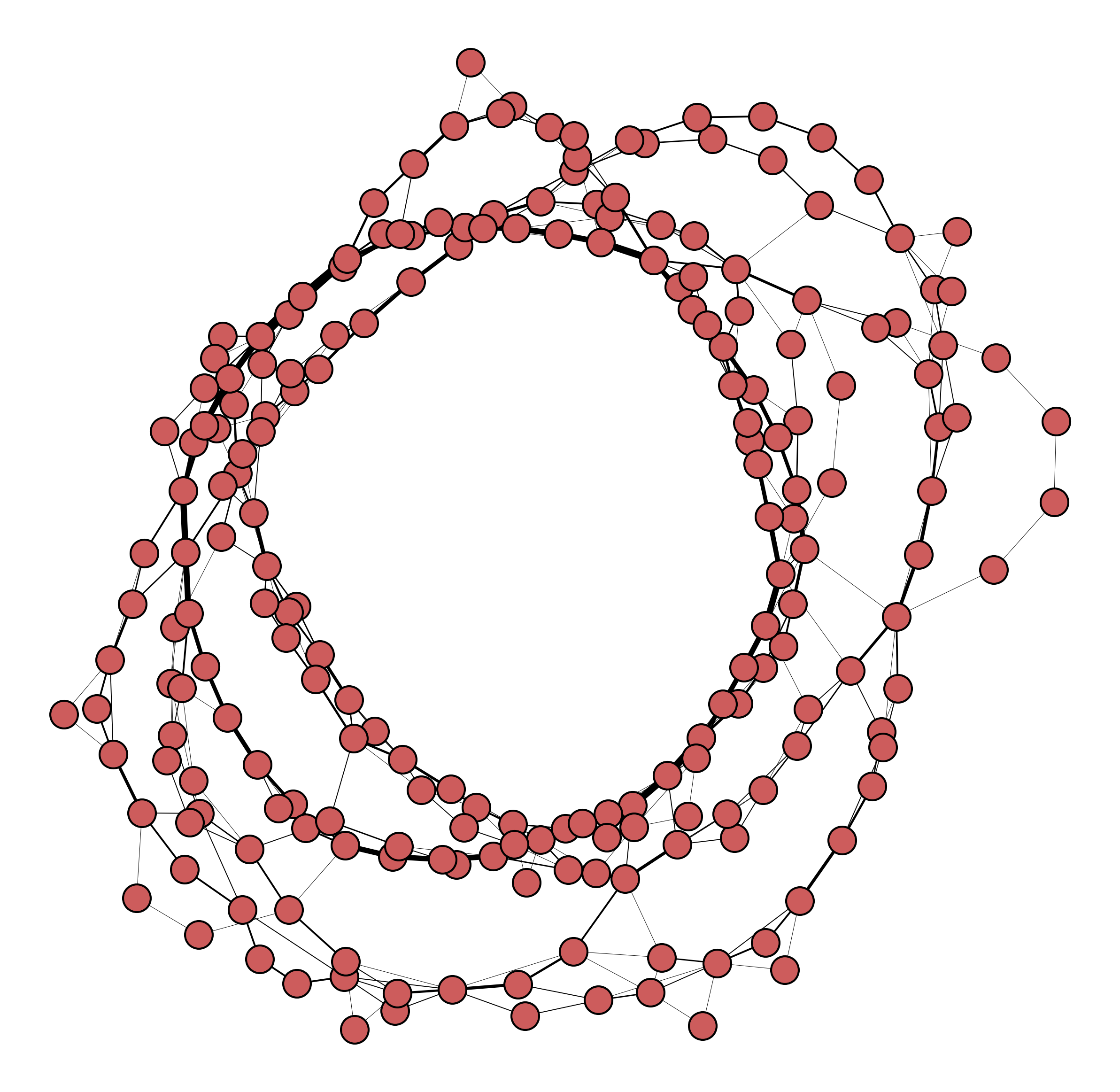} 
        {(e)}
    \end{minipage}
    \begin{minipage}[t]{0.28\textwidth}
        \centering
        \includegraphics[width=\linewidth]{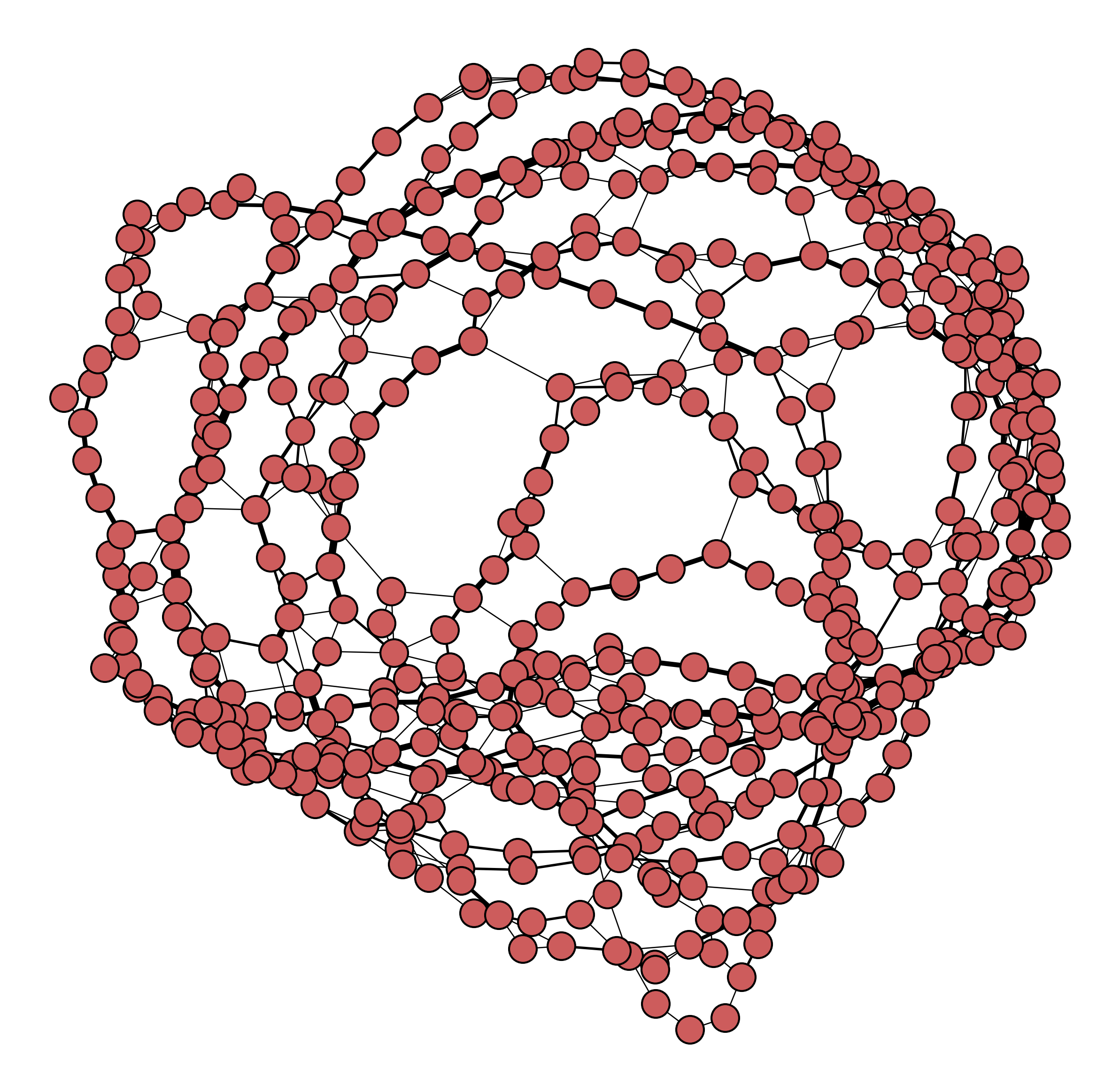} 
        {(f)}
    \end{minipage}
    
    \caption{Transitional complex network topology comparison between OPN and CGSSN for  the $x(t)$ simulation of the Rossler system described in Eq.~\eqref{eq:rossler}. (a) Periodic Rossler Simulation $x(t)$, (b) Periodic OPN ($n=7$) $E'(D_1) = 0.503$., (c) Periodic CGSSN ($n=4$ and $b=12$). $E'(D_1) = 0.026$, (d) Chaotic Rossler Simulation $x(t)$, (e) Chaotic OPN ($n=7$). $E'(D_1) = 0.893$, (f) Chaotic CGSSN ($n=4$ and $b=12$). $E'(D_1) = 0.905$.}
    \label{fig:OPN_vs_CGSSN_Rossler}
\end{figure}

Our first result is from a study of the complex network topology of OPNs compared to CGSSNs. To demonstrate the difference and motivate why the CGSSN outperforms the OPN in terms of dynamic state detection, we use an $x(t)$ simulation of the Rossler system defined as
\begin{equation} \label{eq:rossler}
\frac{dx}{dt}  = -y -z,  \qquad 
\frac{dy}{dt}  = x + ay, \qquad
\frac{dz}{dt}  = b + z(x-c).
\end{equation}
We simulated Eq.~\eqref{eq:rossler} using the \textit{scipy} odeint solver for $t \in [0, 1000]$ with only the last 230 seconds used to avoid transients. The signal was sampled at a rate of $f_s = 22$ Hz.
For periodic dynamics we use system parameters of $[a, b, c] = [0.1, 0.2, 14]$ and for chaotic we set $a=0.15$. 
These simulated signals are shown in Fig.~\ref{fig:OPN_vs_CGSSN_Rossler}. 
To create the OPNs for both signals, we used an embedding delay $\tau = 43$ selected using the multi-scale permutation entropy method and dimension $n=7$. 
The corresponding networks are shown in the second column of Fig.~\ref{fig:OPN_vs_CGSSN_Rossler}. 
To form the CGSSNs we similarly chose $\tau = 43$, but used dimension $n=4$ and $b=12$ for partitioning the SSR with resulting networks shown in the third column.

The resulting OPN and CGSSN from the Rossler system simulations of periodic and chaotic dynamics both capture the increasing complexity of the signal with the dynamic state change. For the periodic signal, the OPN show overarching large loops relating to the periodic nature of the SSR. However, the CGSSN better captures the periodic nature of the trajectory with only a single loop forming. This characteristic of the CGSSN is due to periodic flows never intersecting in the SSR if the signal is sampled at a high enough frequency, there is no or little additive noise, and an appropriately sized delay and dimension are selected. While correctly choosing the delay and dimension is not a trivial task, there is a broad literature on their selection for the SSR task. This work relies on the multi-scale permutation entropy method for selecting the delay and the false-nearest-neighbors algorithm~\cite{Krakovska2015} for selecting an appropriate SSR dimension. However, we found that increasing the dimension one higher than that suggested using false-nearest-neighbors more reliably formed a single loop structure in the CGSSN. Additionally, in Appendix~\ref{app:binning_analysis} we demonstrate that for 23 dynamical systems, setting $b \geq 12$ resulted in only a single loop structure for periodic signals while minimizing the computational demand when using the CGSSN. As such, we set $b=12$ unless otherwise stated.

For the chaotic $x(t)$, the OPN and CGSSN both summarize the topology of the attractor with both networks having a high degree of entanglement with nodes being highly intertwined. This is a typical characteristic of complex transitional networks formed from chaotic signals. Furthermore, it  should  be noted that the CGSSN tends to be more entangled than its OPN counterpart, suggesting that the CGSSN better captures the increase in complexity of the chaotic signal.

To quantify how well the OPN and CGSSN capture the complexity of the signals, we rely on persistent entropy~\cite{Atienza2018}, which was previously adapted~\cite{Myers2019} to study the resulting persistence diagram using the unweighted shortest path distance of complex networks. The normalized persistent entropy~\cite{Atienza2020,Atienza2021} is defined as 
\begin{equation} \label{eq:persistent_entropy}
E'(D)=\frac{- \sum_{x \in D} \frac{\pers(x)}{\mathscr{L}(D)}\log_2\left(\frac{\pers(x)}{\mathscr{L}(D)}\right)}{\log_2\big(\mathscr{L}(D))},
\end{equation}
where $\mathscr{L}(D) = \sum_{x \in D} \pers(x)$ with $\pers(x) = |b-d|$ as the lifetime or persistence of point $x = (b,d)$ in a persistence diagram $D$. For studying the complexity of transitional network we apply this score to the one-dimensional persistent diagram $D_1$, which measures the loop structures in the network.
This score yields a value close to zero for networks with a single loop structure corresponding to periodic dynamics and a value close to one for chaotic dynamics with highly intertwined networks.
For our example OPN and CGSSNs in Fig.~\ref{fig:OPN_vs_CGSSN_Rossler} we get normalized persistent entropy scores of $0.503$ and $0.893$ for periodic and chaotic OPNs, respectively, and $0.026$ and $0.905$ for CGSSNs. These statistics show that the CGSSN outperforms the OPN with a significantly larger difference in the entropy values. This is mainly due to the CGSSN having a score near zero for periodic dynamics due to its general loop structure compared to the periodic OPN having several loops.
This result comparing the OPN and CGSSN suggests that the CGSSN will outperform the OPN for the dynamic state detection task. 
With this single case under our belt, we turn our attention to an empirical study of this characteristic over more dynamical systems. 

\subsection{Empirical Testing of Dynamic State Detection for 23 Continuous Dynamical Systems} \label{ssec:dynamic_state_detection_23_systems}

The previous example in Section~\ref{ssec:dynamic_state_detection_results_rossler} showed the improved dynamic state detection performance of the CGSSN over the OPN for a single example (R\"{o}ssler System). However, we want to show that this improvement is present over various systems. To do this, we use 23 continuous dynamical systems listed in the Appendix~\ref{app:data} with details on the simulation method--each system was simulated for both periodic and chaotic dynamics.

For each periodic and chaotic signal, we calculate the resulting persistence diagram of the OPN and CGSSN using each of the distance methods (unweighted shortest path, shortest weighted path, weighted shortest path, and diffusion distance). 
We then compare the collection of persistence diagrams for a specific network type (OPN or CGSSN) and distance measure by calculating the bottleneck distance matrix between each persistence diagram. 
The bottleneck distance $d_{BN}(D, F)$ is a similarity measure between two persistence diagrams ($D$ and $F$). It is calculated as the sup norm distance between the persistence diagrams, where the persistence diagrams are optimally matched with the distance between matched persistence points being at most $d_{BN}$. 
The bottleneck distance matrix $\mathbf{D}_{BN}$ is calculated by finding $d_{BN}$ between all persistence diagrams.

The question we are trying to answer is if periodic and chaotic dynamics result in similar persistence diagrams across multiple systems. 
To answer this, we first use a lower-dimensional projection of $\mathbf{D}_{BN}$ by implementing the Multi-Dimensional Scaling (MDS) projection to two dimensions. 
To measure how well the dynamics delineate on the MDS projection, we use a Support Vector Machine (SVM) with a Radial Basis Function (RBF).
Note that because the MDS does not allow for the mapping of previously unseen points, we cannot use this procedure for a proper classification test as we cannot approximate training error.  
However, we can use this procedure to see if the the persistence diagrams of different classes are separated with respect to the bottleneck distance.

We fit the SVM using the default \texttt{SKLearn} SVM parameters package.
The resulting separations for periodic and chaotic dynamics using the OPN (left) and CGSSN (right) are shown in Appendix~\ref{app:SVM_shortest_path_distances}. 
These separations are for the diffusion distance calculation as it provided the best results for both the OPN and CGSSN. 
However, we also include similar figures for other choices of distances in Appendix~\ref{app:SVM_shortest_path_distances}. 

\begin{figure}
    \centering
    \begin{minipage}[t]{0.43\textwidth}
        \centering
        {Diffusion distance of OPN}
        \includegraphics[width=\linewidth]{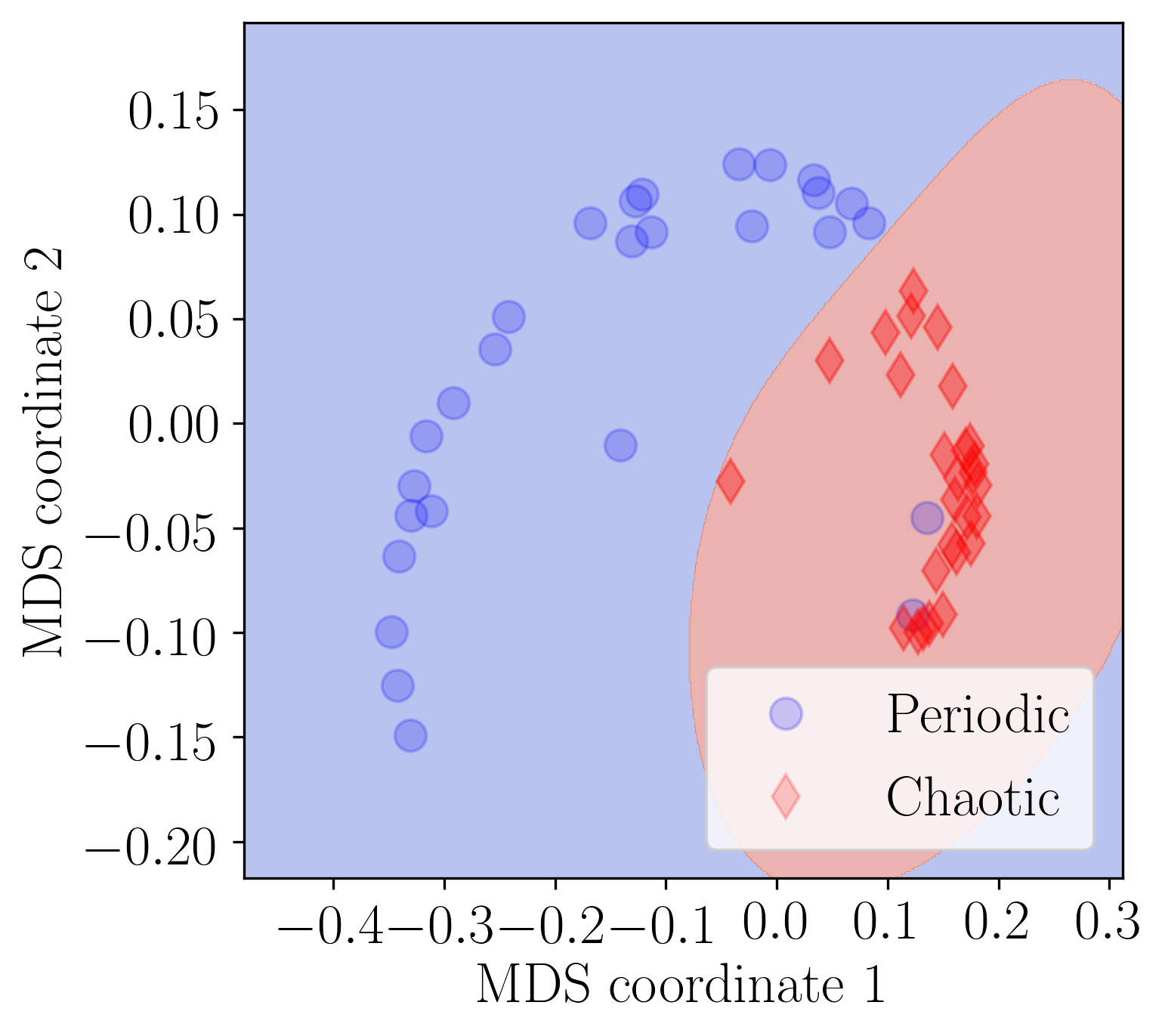} 
    \end{minipage}
    \hspace{10mm}
    \begin{minipage}[t]{0.4\textwidth}
        \centering
        {Diffusion distance of CGSSN}
        \includegraphics[width=\linewidth]{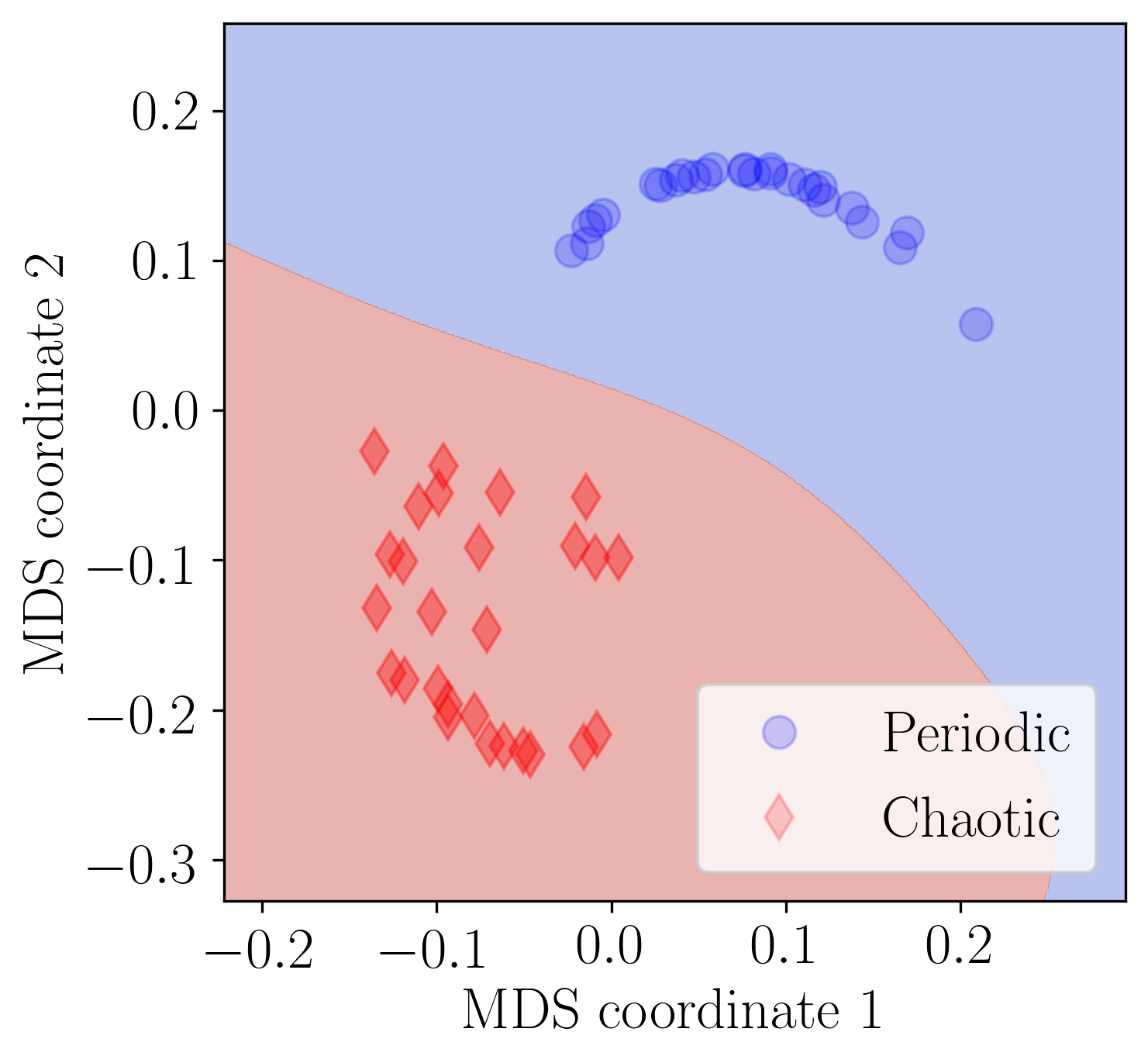} 
    \end{minipage}
    
    \caption{Two-dimensional MDS projection 
    of the bottleneck distances between persistence diagrams of the chaotic and periodic dynamics with an SVM radial bias function kernel separation. This separation analysis was repeated for the OPNs and CGSSNs using the diffusion distance.}
    \label{fig:MDS_diffusion_distaance_OPN_vs_CGSSN}
\end{figure}

Figure~\ref{fig:MDS_diffusion_distaance_OPN_vs_CGSSN} demonstrates the significant improvement in dynamic state detection of the CGSSN over the OPN. This is shown with the periodic and chaotic networks being clustered for the CGSSN (right of Fig.~\ref{fig:MDS_diffusion_distaance_OPN_vs_CGSSN}) with no overlap compared to the OPN (left of Fig.~\ref{fig:MDS_diffusion_distaance_OPN_vs_CGSSN}) having some overlap between periodic and chaotic dynamics. This is further shown with the SVM kernel being able to separate the periodic and chaotic regions for the CGSSN easily.
To better compare all distance measures and complex network combinations, we quantify the performance of each SVM kernel using the accuracy of the separation. We repeated this accuracy calculation 100 times for each combination using 100 random seeds to generate the SVM kernels. The resulting average accuracies with standard deviation uncertainties are reported in Table~\ref{tab:accuracies}.
\begin{table}
\centering
\caption{Accuracies for SVM seperation of MDS projections for dynamic state detection. Uncertainties are recorded as one standard deviation for random seeds 1-100.}
\label{tab:accuracies}
\begin{tabular}{cccc}
\textbf{Network} & \textbf{Distance} & \textbf{Average Separation Accuracy} & \textbf{Uncertainty} \\ \hline
OPN & Shortest Unweighted Path Distance & 80.7\% & 1.5\% \\
OPN & Shortest Weighted Path Distance & 88.9\% & 0.0\% \\
OPN & Weighted Shortest Path Distance & 88.9\% & 0.0\% \\
OPN & Lazy Diffusion Distance & 95.0\% & 0.9\% \\
CGSSN & Shortest Unweighted Path Distance & 98.1\% & 0.9\% \\
CGSSN & Shortest Weighted Path Distance & 100.0\% & 0.0\% \\
CGSSN & Weighted Shortest Path Distance & 98.1\% & 0.9\% \\
CGSSN & Lazy Diffusion Distance & 100.0\% & 0.0\% \\ \hline
\end{tabular}
\end{table}

Based on the results in Table~\ref{tab:accuracies}, the CGSSN outperforms the OPN for all distance measures. 
Additionally, we found 100\% separation accuracy for both the shortest weighted path and diffusion distances when combined with the CGSSN. 
We believe this performance improvement is due to the coarse-graining procedure capturing the SSR vector's amplitude information which is discarded when identifying permutations in the OPN.

\subsubsection{$n$-Periodic Systems}
\begin{figure}%
    \centering
    \includegraphics[width=\textwidth]{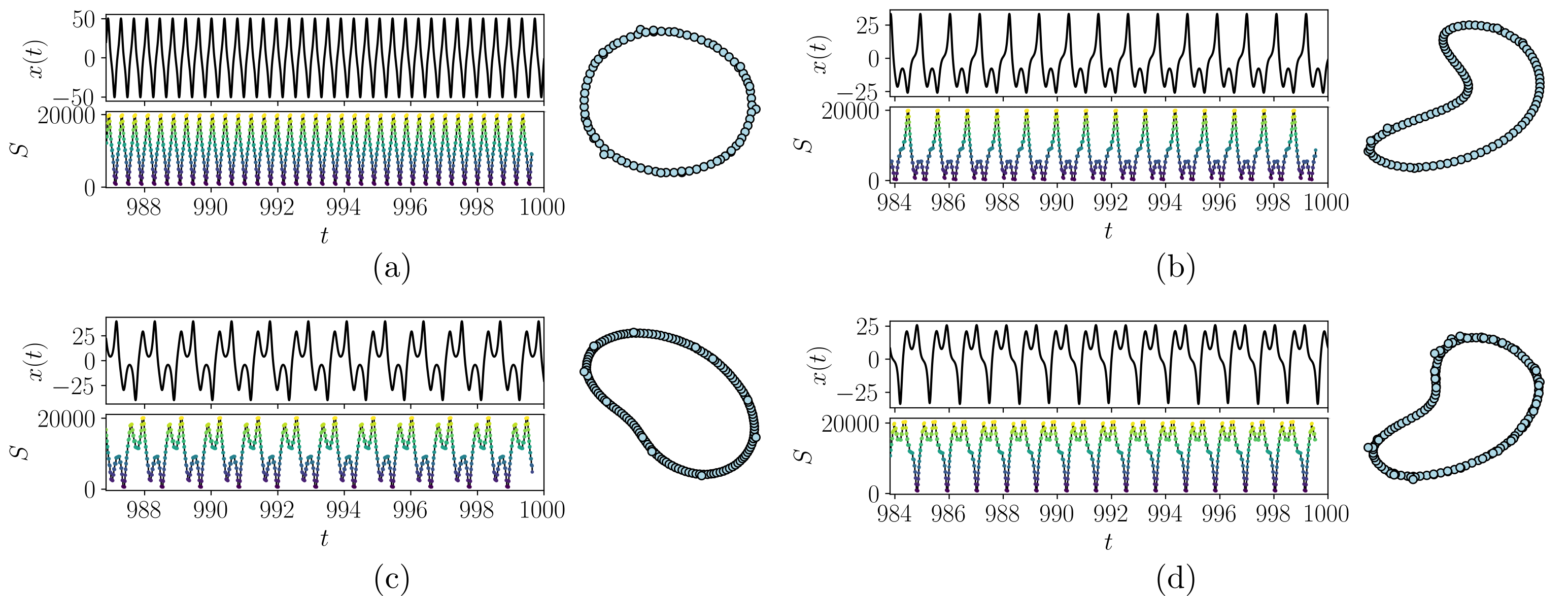}
    \caption{CGSSN results for four multi-periodic cases of the Lorenz system. The sequence of A's and B's below each image indicates the oribital sequence around the attractors A and B in the system. (a) AB $\rho=350$, (b) AAB $\rho=100.5$, (c) AABB $\rho=160$, (d) ABBABB $\rho=99.65$. As expected, all four cases result in a single loop CGSSN. These networks were generated using $n=4$ and $b=12$.}
    \label{fig:n-periodic}
\end{figure}

Based on the state space embedding structure of a system, one may expect that for a 2 or 3-periodic system that the CGSSN may result in 2 and 3 loops respectively, but this is not the case. 
In general, for an $n$-periodic system, we expect the CGSSN to contain only a single loop, and so we caution the user that this method will likely not be able to differentiate differences in the periodicity. 
We demonstrate this nuance by showing CGSSN results on the Lorenz system for multi-periodic responses. Fig.~\ref{fig:n-periodic} shows the corresponding CGSSNs for the Lorenz system varying the $\rho$ parameter to obtain multi-periodic responses. 
The networks are labeled with a sequence of $A$'s and $B$'s where each letter corresponds to a loop in the trajectory around one of the attractors. 
For example $AAB$ trajectory would be two loops around A and one around B before repeating the cycle. 
For all four cases shown, a single loop is obtained in the CGSSN even though the system exhibits multi-periodicity. 

\subsection{A Remark on Discrete Maps}
\label{ssec:discrete_maps}
\begin{figure}
    \centering
    \begin{minipage}[t]{\textwidth}
        \includegraphics[width=0.9\textwidth]{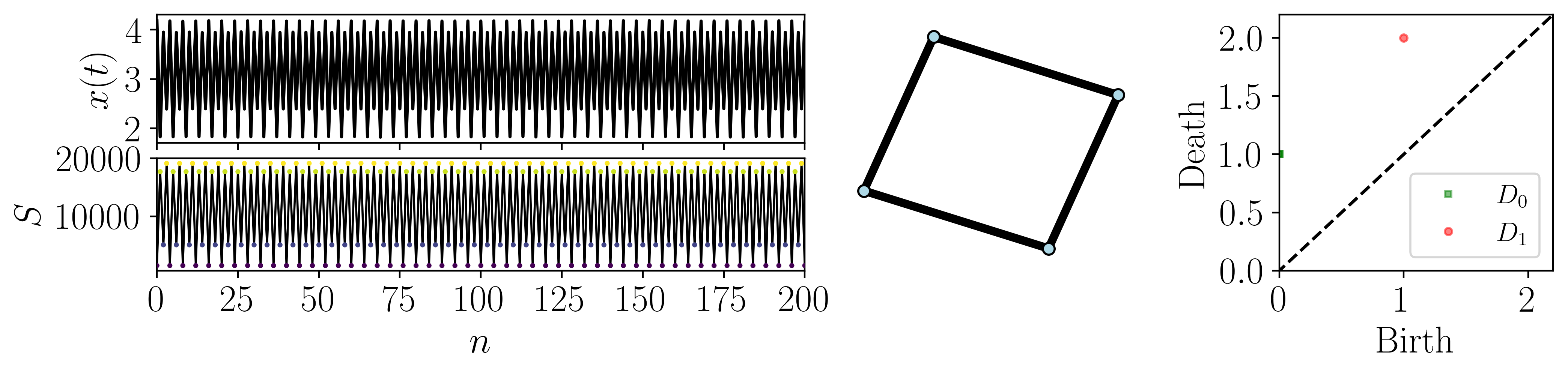}
    \end{minipage}
    \begin{minipage}[t]{\textwidth}
        \includegraphics[width=0.9\textwidth]{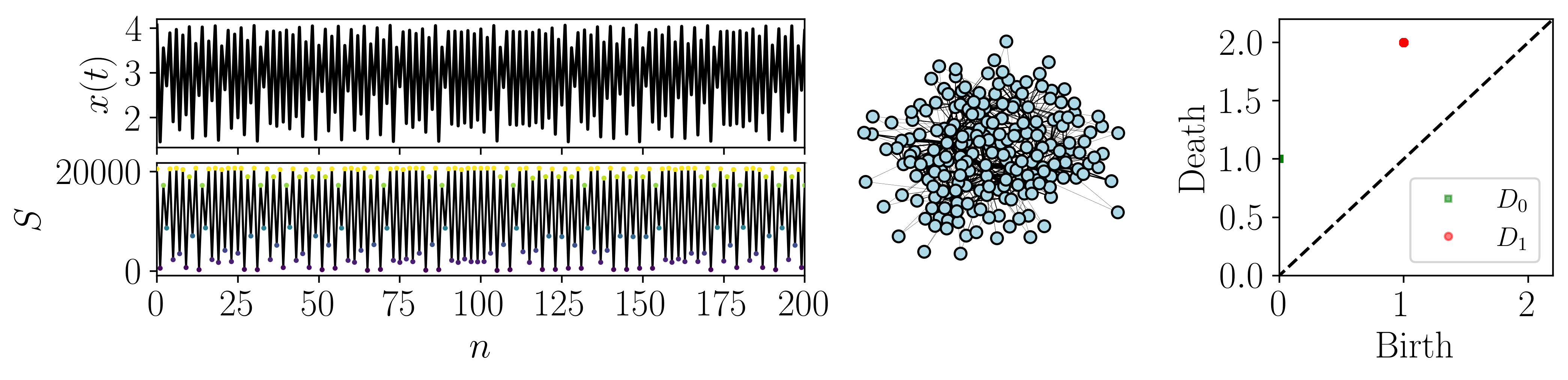}
    \end{minipage}
    \caption{CGSSN results for the periodic (top row) and chaotic (bottom row) logistic map using the unweighted shortest path. The system responses are shown on the left along with the permutation sequence. The network representations are in the middle with the persistence diagrams on the right. Both networks exhibit the same persistence diagram due to the limited possible system states for the periodic case.}
    \label{fig:logistic_results}
\end{figure}

As we demonstrated in Section.~\ref{ssec:dynamic_state_detection_23_systems}, the CGSSN method allows for efficient and accurate dynamic state detection over a range of continuous dynamical systems. Discrete maps are another subset of dynamical systems where it would be useful to apply these tools; however, care must be taken for this type of system to ensure that the CGSSN is a suitable approach. This is because in discrete systems, there are typically far fewer states that the system can exhibit so in some cases the CGSSN may not contain any loops, but the response is still periodic leading to an incorrect classification in the model. To demonstrate, we show the CGSSNs for the periodic and chaotic logistic map in Fig.~\ref{fig:logistic_results} where the unweighted shortest path distance was used to compute persistence. We see that the CGSSNs show vastly different structures where the periodic network contains a single loop and the chaotic network is tangled. However, the persistence diagrams for these networks appear to be equivalent because the networks were unweighted and all of the loops in the chaotic network are exactly the same size as the periodic case. Due to only having 4 possible states in the periodic logistic map here, the network loop does not provide enough of a difference to automatically classify it as either dynamic state. We note that the chaotic persistence diagram contains more loops than the periodic case here, but all are the same persistence lifetime. In the case of a continuous system where many more states are possible, these loops will be larger in size and the persistence diagram will reflect those differences allowing for classification of the dynamic state. In this case, when other distances are used such as the shortest weighted path, the resulting persistence diagrams have the forms that we expect for periodic and chaotic behaviors due to the weighting of the edges influencing the persistence lifetime of that loop. 

\begin{figure}
    \centering
    \begin{minipage}[t]{\textwidth}
        \includegraphics[width=0.9\textwidth]{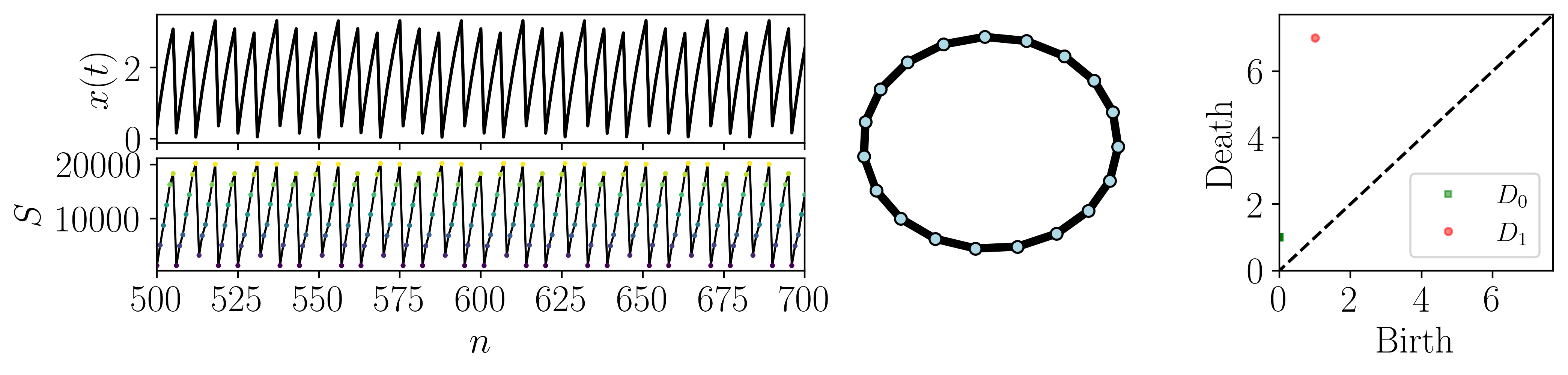}
    \end{minipage}
    \begin{minipage}[t]{\textwidth}
        \includegraphics[width=0.9\textwidth]{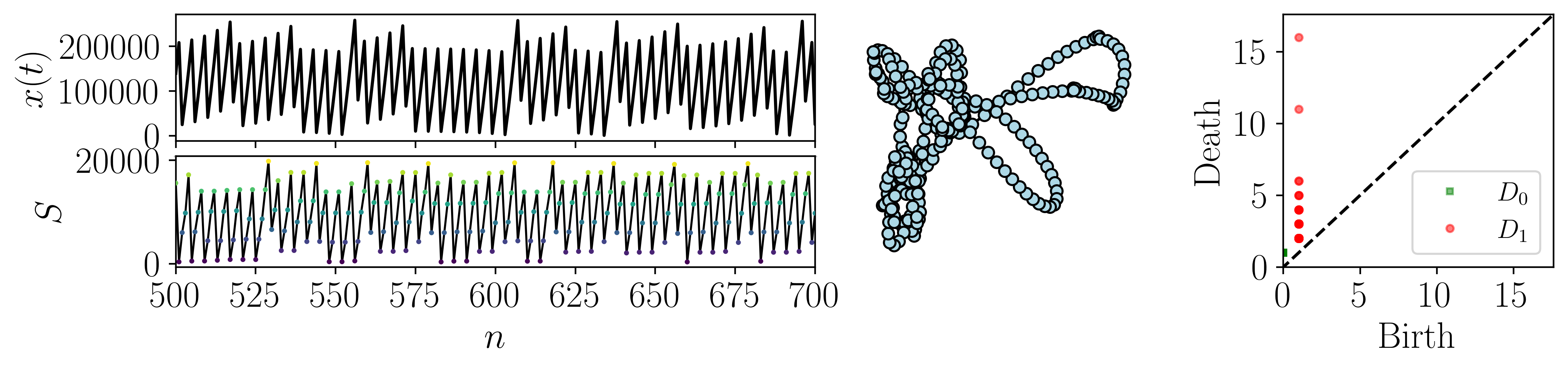}
    \end{minipage}
    \caption{CGSSN results for the periodic (top row) and chaotic (bottom row) linear congruential generator map using the unweighted shortest path. The system responses are shown on the left along with the permutation sequence. The network representations are in the middle with the persistence diagrams on the right. Both networks exhibit the distinct persistence diagram structures due to the larger loop in the periodic network.}
    \label{fig:lcgm_results}
\end{figure}

In the case where the system being studied can exhibit many possible states in its periodic response, a single loop will form the CGSSN and the persistence diagram will show a persistence pair with a long lifetime. For example, we demonstrate this behavior on the 3 periodic linear congruential generator map in Fig.~\ref{fig:lcgm_results}. The results in Figs.~\ref{fig:logistic_results} and \ref{fig:lcgm_results} demonstrate that this method should be used with caution on discrete systems and for systems with enough states that approach the behavior of a continuous system, the CGSSN persistence diagrams can provide a correct dynamic state detection.

\subsection{Noise Sensitivity} \label{ssec:noise_sensitivity}

\begin{figure}
  	\begin{center}
    	\includegraphics[width=.5\textwidth]{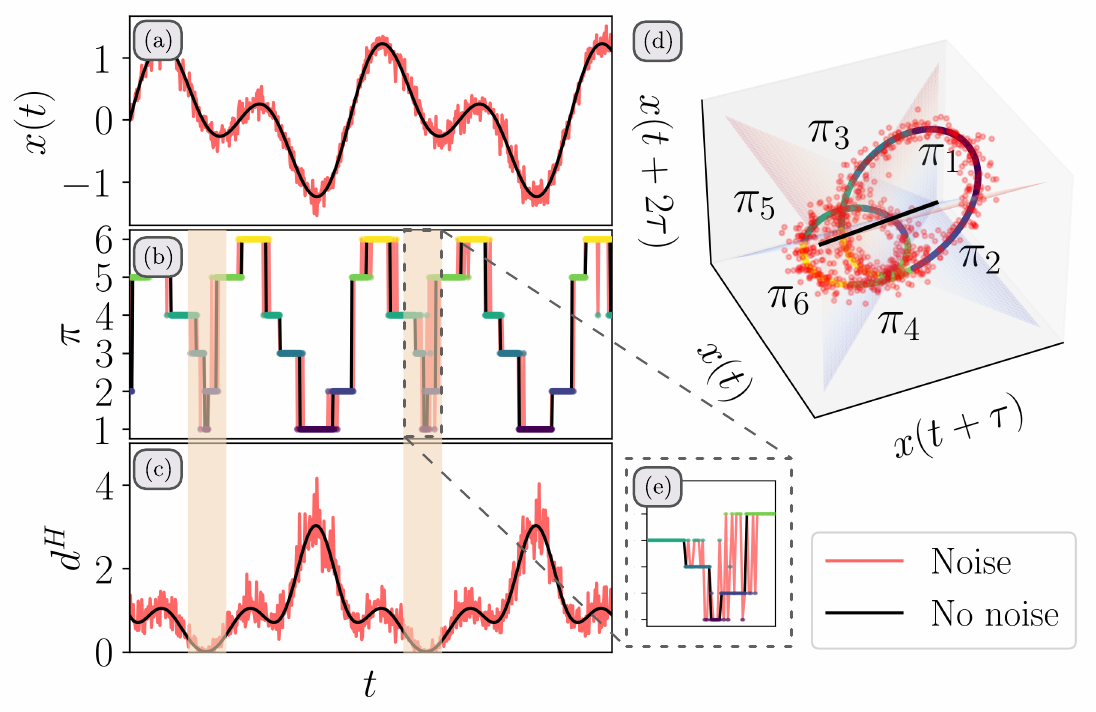}
 	\end{center}
	\caption{The three-dimensional state space reconstruction (d) from the signal $x(t)$ with and without additive noise (a) shows as the distance to the hyperdiagonal $d^H$ (c) becomes small, undesired permutation transitions (b)---with zoomed-in section shown in (e)---occur as shown in the orange highlighted regions.}
  	\label{fig:hyperdiagonal_distance_and_permutations_issue}
\end{figure}

One issue with ordinal partition networks is they are not exceptionally resilient to noise.
Indeed, one can think of the ordinal partition network as being the 1-skeleton of the nerve of a particular closed cover of the state space, delineated by the hyperplanes $x_i \leq x_j$.
Consequently, when noise is injected into the system, there are superfluous transitions when nearing one of these boundaries.
For example, consider the signal and its embedding into $\mathbb{R}^3$ in Fig.~\ref{fig:hyperdiagonal_distance_and_permutations_issue}.

\begin{figure}
  \centering
  \includegraphics[width=.95\textwidth]{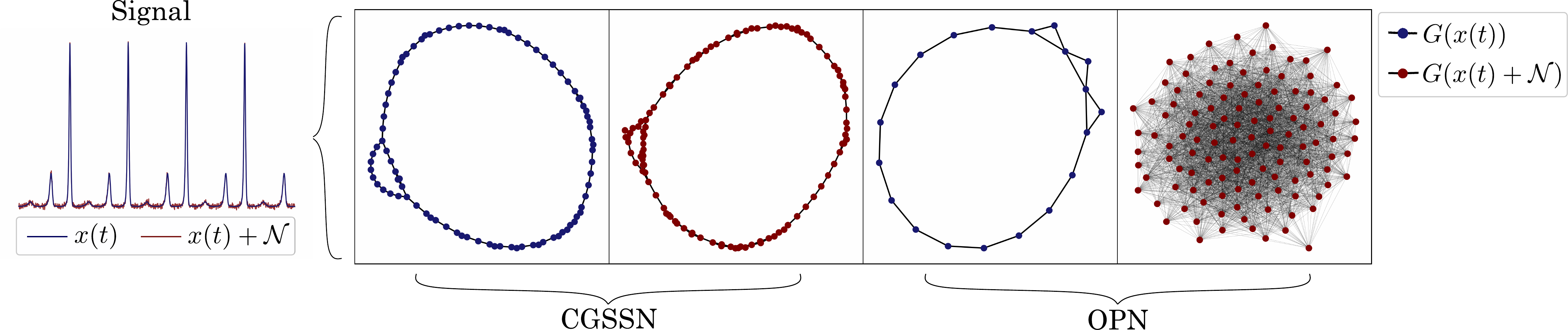}
  \caption{Example demonstrating the importance of choosing an appropriate network formation method when there is additive noise in the signal. The CGSSN retains the graph structure even with additive noise; in contrast, the OPN network loses all resemblance to the noise-free topological structure even with a small amount of additive noise. $x(t)$ is the signal, $\mathcal{N}$ is additive noise, and $G(x)$ is the graph representation of $x$.}
  \label{fig:network_type_consideration_figure_for_noise}
\end{figure}

This effect becomes even more prominent near an intersection of multiple hyperplanes.
As the distance to the hyperdiagonal $d^H$ becomes small, we see a significant increase in seemingly superfluous transitions between permutations $\pi$ (highlighted in orange in Fig.~\ref{fig:hyperdiagonal_distance_and_permutations_issue}).
This issue is even more exaggerated when the embedded signal is consistently close to the hyperdiagonal, which results in network representations whose shape carries no information on the underlying dynamical system (e.g., see the signal and far-right OPN in Fig.~\ref{fig:network_type_consideration_figure_for_noise}).
This is particularly detrimental when we attempt to include the weighting information, as the flips can skew the count for the number of times a boundary is crossed.

Certain network representations of time series are naturally more noise-robust than others.
For example, Fig.~\ref{fig:network_type_consideration_figure_for_noise} shows the OPN and CGSSN for the signal with and without noise.
This example demonstrates that the CGSSN is the best choice for this signal with only minor changes in its shape, while the OPN loses all resemblance to the noise-free network.

\begin{figure}
    \centering
    \begin{minipage}[t]{0.48\textwidth}
        \centering
        {OPN}
        \includegraphics[width=\linewidth]{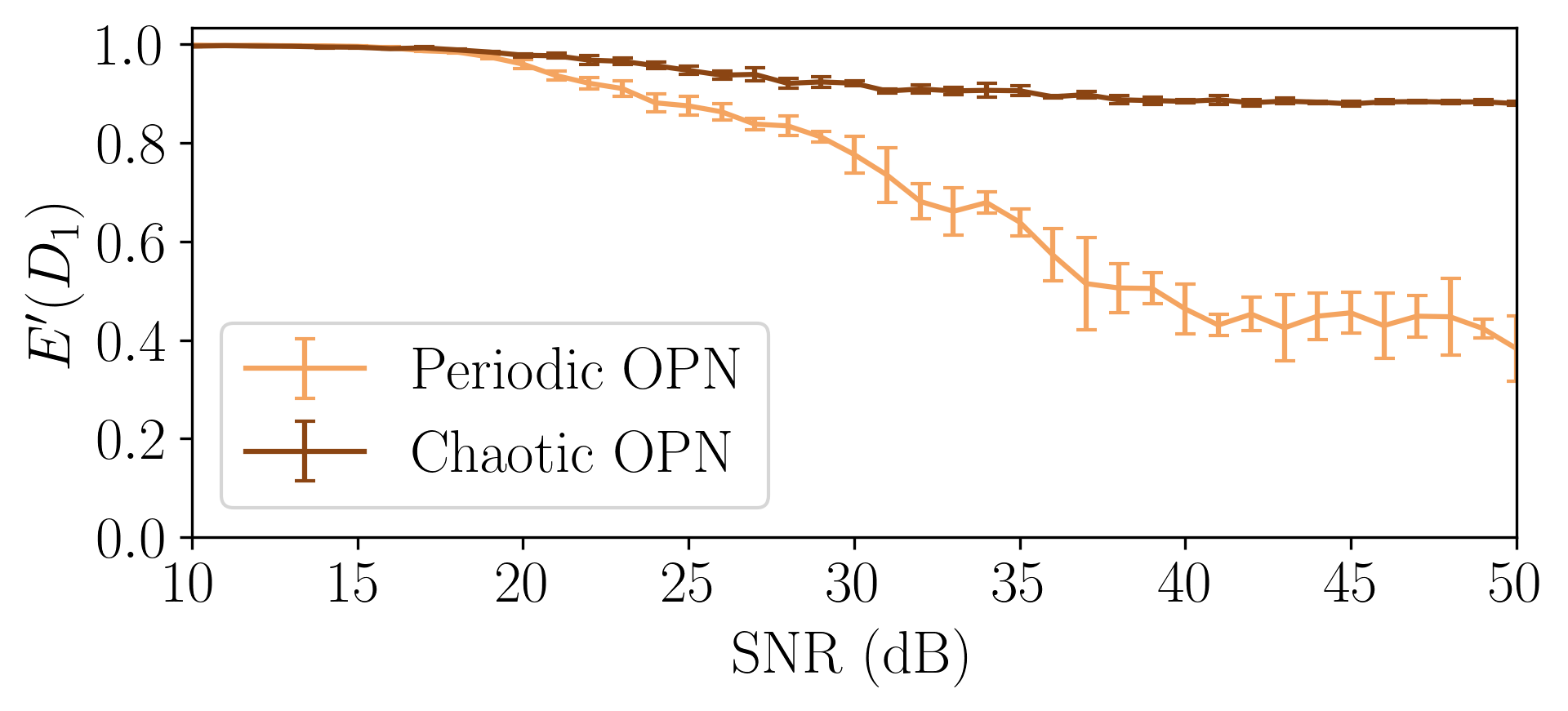}
    \end{minipage}
    \hspace{3mm}
    \begin{minipage}[t]{0.48\textwidth}
        \centering
        {CGSSN}
        \includegraphics[width=\linewidth]{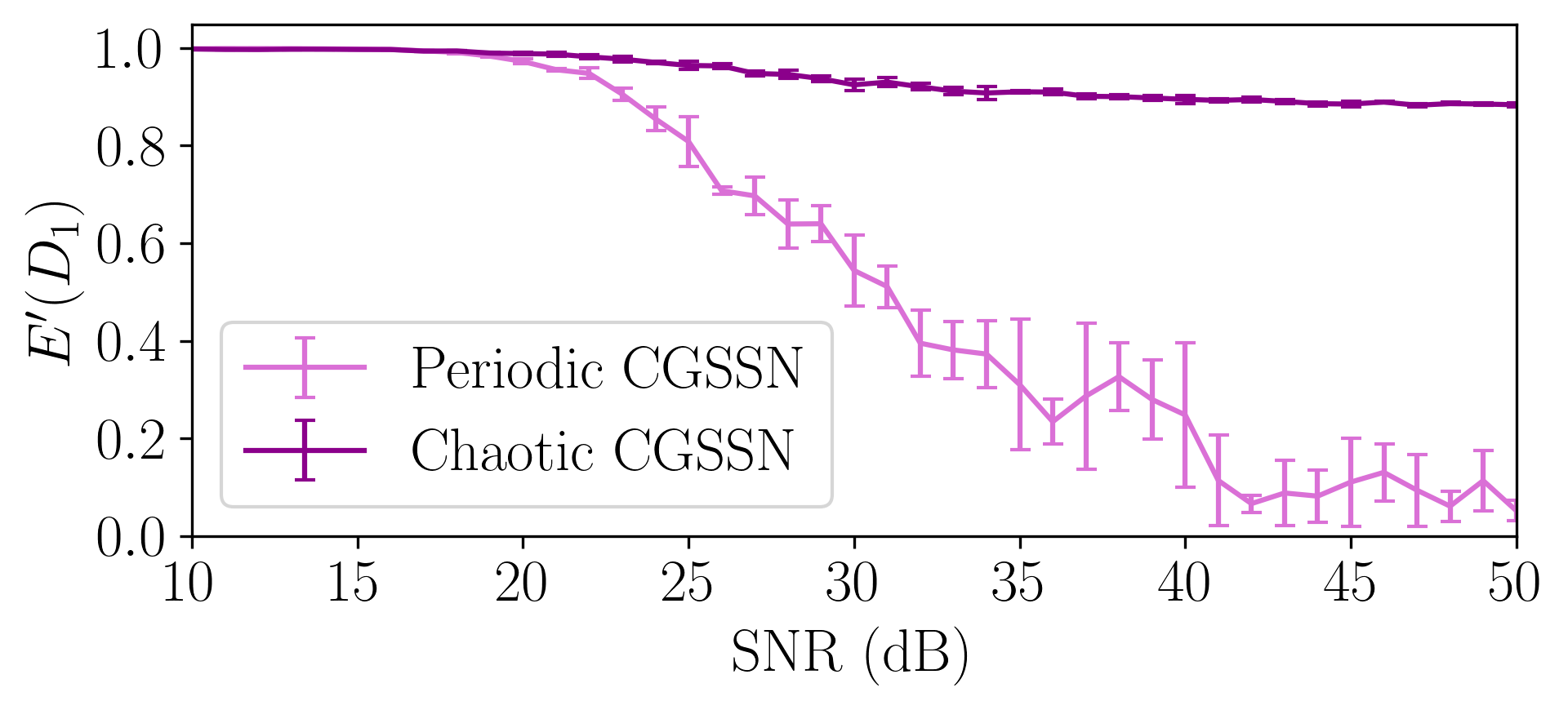} 
    \end{minipage}
    
    \caption{Noise robustness analysis of dynamic state detection using the summary statistic persistent entropy (see Eq.~\eqref{eq:persistent_entropy}) for OPN and CGSSN with increasing SNR on a periodic Rossler simulation from Eq.~\eqref{eq:rossler}.
    }
    \label{fig:noise_robustness_opn_cgssn_persistent_entropy}
\end{figure}

Outside of this sensitivity to the hyperdiagonal, we also found that the CGSSN is more noise robust than the OPN for other signals. For example, in Fig.~\ref{fig:noise_robustness_opn_cgssn_persistent_entropy} we show the normalized persistent entropy statistic from Eq.~\eqref{eq:persistent_entropy} calculated for the periodic and chaotic simulations of the R\"{o}ssler system defined in Eq.~\eqref{eq:rossler} when additive noise is present in the signal. We incremented the additive noise using the Signal-to-Noise Ratio (SNR). 
The SNR (units of decibels) is defined as
${\rm SNR} = 20 \log_{10}(A_{\rm signal}/A_{\rm noise})$, 
where $A_{\rm signal}$ and $A_{\rm noise}$ are the root-mean-square amplitudes of the signal and additive noise, respectively. 
This result shows that for this signal the OPN network is only robust down to an SNR of approximately 32 dB of additive Gaussian noise, while the CGSSN is able to separate periodic from chaotic dynamics down to approximately 23 dB. We found similar results for the other 22 dynamical systems investigated in this work.

\subsection{Experimental Results}
\label{ssec:experimental}
\begin{figure}[htbp]
    \centering
    \includegraphics[width=0.9\textwidth]{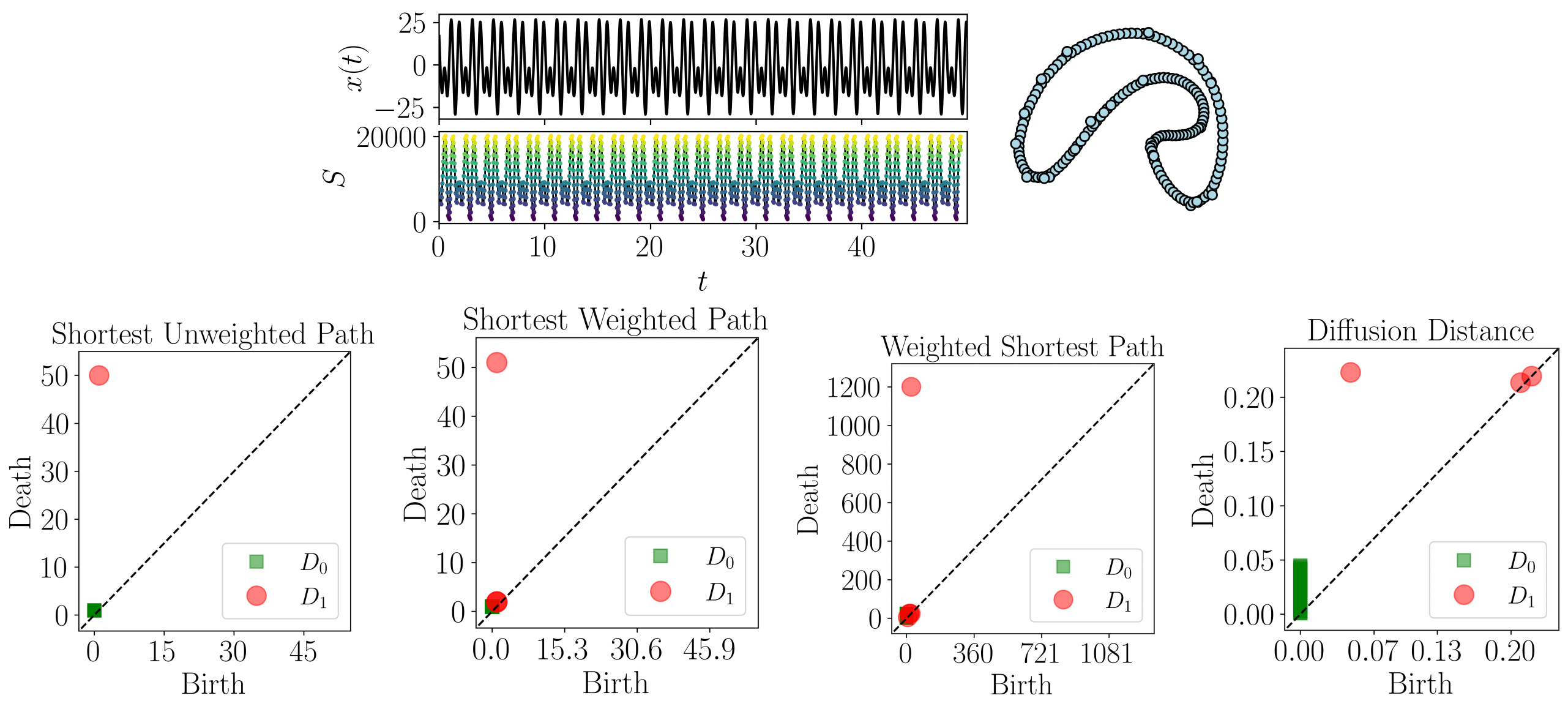}
    \caption{CGSSN results for the forced single magnetic pendulum under conditions that yield a periodic response. The top left images show the time series and permutation sequence and the top right shows the coarse grained state space network. The bottom row shows the corresponding persistence diagrams for the network under the distance metric in the title of each diagram.}
    \label{fig:magnetic_periodic}
\end{figure}

\begin{figure}[htbp]
    \centering
    \includegraphics[width=0.9\textwidth]{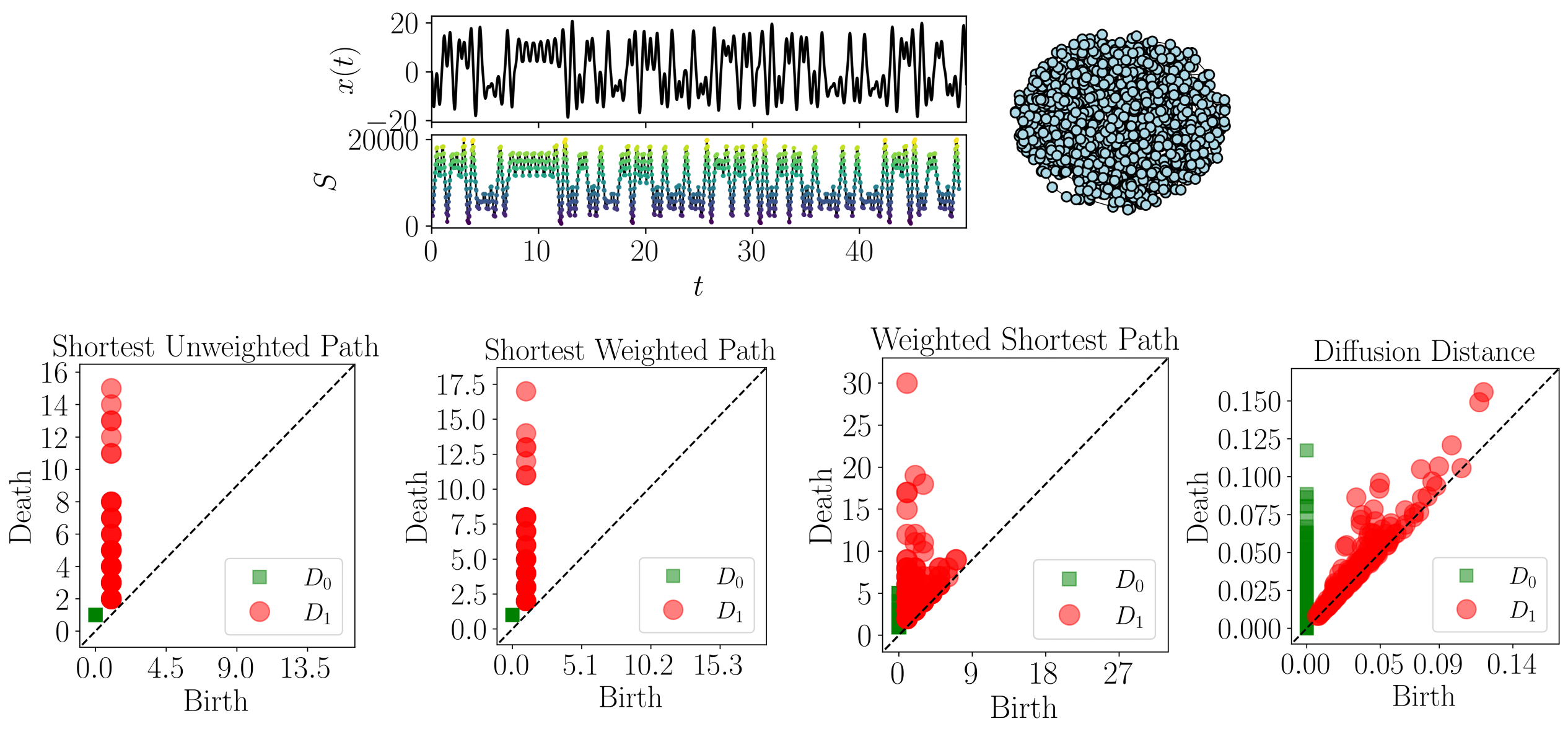}
    \caption{CGSSN results for the forced single magnetic pendulum under conditions that yield a chaotic response. The top left images show the time series and permutation sequence and the top right shows the coarse grained state space network. The bottom row shows the corresponding persistence diagrams for the network under the distance metric in the title of each diagram.}
    \label{fig:magnetic_chaotic}
\end{figure}

To validate these tools, we apply them to experimental data collected from a base excited magnetic pendulum \cite{Myers_2020_pendulum}. This system was shown to exhibit periodic and chaotic behavior under different parameters and the CGSSN persistence diagrams were generated for each case using all 4 distance measures presented in this paper. Figure \ref{fig:magnetic_periodic} shows the corresponding time series, permutation sequence, CGSSN, and persistence diagrams for the periodic response. We see that for all of the distance metrics, there is a clear singular cycle that forms with a significant persistence lifetime. Conversely, the same results are presented for the chaotic response in Fig.~\ref{fig:magnetic_chaotic} where we see a drastically different distribution of persistence pairs corresponding to the high number of cycles present in the chaotic CGSSN. The results presented here are in agreement with our work in \cite{Myers_2020_pendulum}.


\section{Conclusion} \label{sec:conclusion}

In this work, we developed a novel framework for studying CGSSNs using persistent homology. 
We showed that the CGSSN outperformed the standard ordinal partition network in both noise robustness and dynamic state detection performance, with the CGSSN reaching 100\% separation accuracy for dynamic state detection for 23 continuous dynamical systems. 
This is in comparison to the OPN, which could at most reach 95\% accuracy. This approach was validated using data from a magnetic pendulum experiment to show that the topological structure for periodic and chaotic timseries are captured in the resulting persistence diagrams.  

In this work, we only investigated the most straightforward construction of the CGSSN. 
Namely, the equal-sized hyper-cube tessellation cover of the SSR domain. 
Possible improvements to the CGSSN could be through a data-dependent adaptive cover algorithm.
We also suspect that other choices of distances could provide improvements for the given pipeline. 

Another future direction would be to prove a stability theorem for the CGSSN. 
That is, can we show that for a noisy version of a signal, the resulting CGSSN, and subsequently the computed persistence diagram, is similar to the ground truth. 
It would also be interesting to study how the CGSSN could serve to detect quasiperiodicity. 
We believe that the torus shape associated to the SSR of quasiperiodic signals could be captured using the CGSSN as it accounts for the signal amplitude.
\section*{Acknowledgements}
This material is based upon work supported by the Air Force Office of Scientific Research under award number FA9550-22-1-0007.

\section*{References}
\bibliographystyle{ieeetr}
\bibliography{./CGSSN_bib}

\newpage
\appendix
\section{Coarse Graining Size Analysis} \label{app:binning_analysis}
\begin{figure}
    \includegraphics[width = 0.8\textwidth]{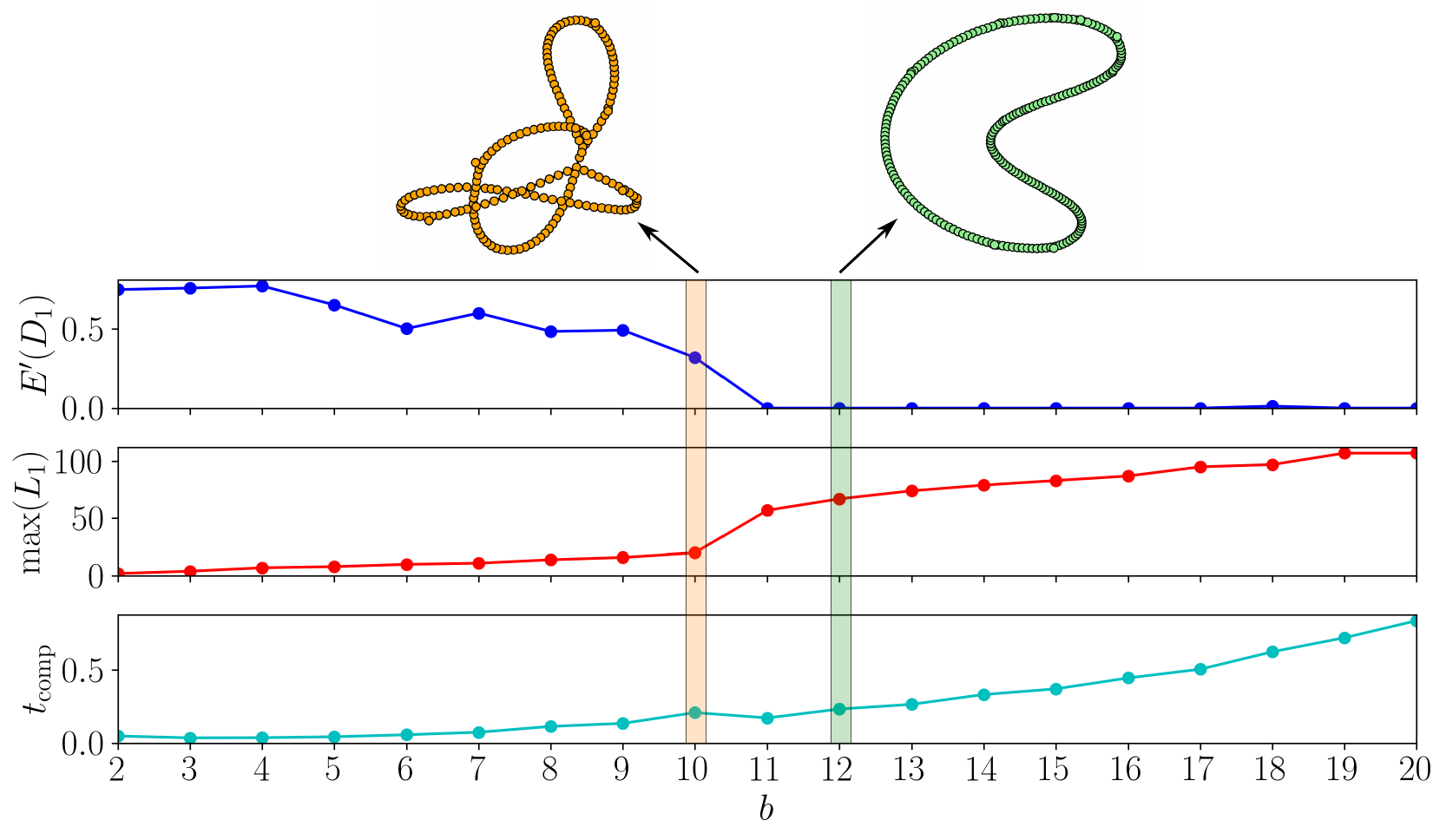}
	\caption{Normalized persistent entropy $E'(D_1)$, the maximum lifetime $\max(L_1)$, and computation time $t_{\rm comp}$ for the CGSSN formed with dimension $n=4$ and $b \in [2, 20]$ for the Rossler system in Eq.~\eqref{eq:rossler} with example CGSSNs shown at $b=10$ and $b=12$.}
	\label{fig:binning_analysis_rossler}
\end{figure}

To determine the optimal binning size we we investigate how the structure of the resulting CGSSN changes as more states are used with $b$ increasing. We considered $b \in [2, 20]$ as more than 20 bins per dimension becomes computationally expensive without increasing the performance (see Fig.~\ref{fig:binning_analysis_rossler}). To summarize the shape of the network we use the maximum lifetime of one-dimensional features (loops) as $\max(L_1)$ and the normalized persistence entropy $E'(D_1)$ defined in Eq.~\eqref{eq:persistent_entropy} using the shortest unweighted path distance. The goal is to find a fine enough granularity (large enough $b$) that a periodic, noise-free signal will create a signal loop structure in the CGSSN. This loop structure should result with a persistent entropy of approximately zero. The idea behind this is based on a periodic attractor's SSR never intersecting if a suitably high dimension is selected. 

We point the reader to our work in \cite{Myers2020a} for a comprehensive analysis to choosing a suitable embedding dimension for the problem. It was found that dimensions of $n=4~\mathrm{or}~5$ are suitable for most continuous systems. For the 23 dynamical systems selected a dimension $n=4$ is greater than the dimension of the attractor and will be used unless otherwise stated. Let us first investigate a suitable number of bins $b$ for the Rossler system defined in Eq.~\eqref{eq:rossler} with the $E'(D_1)$, $\max(L_1)$, and computation time $t_{\rm comp}$ calculated as $b$ is increased from 2 to 20 shown in Fig.~\ref{fig:binning_analysis_rossler}.
This result show a sudden drop in $E'(D_1)$ and increase in $\max(L_1)$ from going from 10 to 11 bins. This is due the the granularity of the coarse-graining procedure being fine enough that the hypercubes do not capture multiple segments of the periodic flow. This is shown with the two CGSSNs at $b=10$ and $b=12$ where at $b=10$ we have multiple intersections of the network while at $b=12$ there are no intersections and we only have a single loop structure. Another characteristic is the exponentially increasing computation time $t_{\rm comp}$ as $b$ increases. As such, we want to optimize the choice of $b$ to capture the necessary complexity of the attractor while also minimizing the computation time. For this example a suitable $b=12$ would be the best choice. 

\begin{figure}
    \includegraphics[width = 0.8\textwidth]{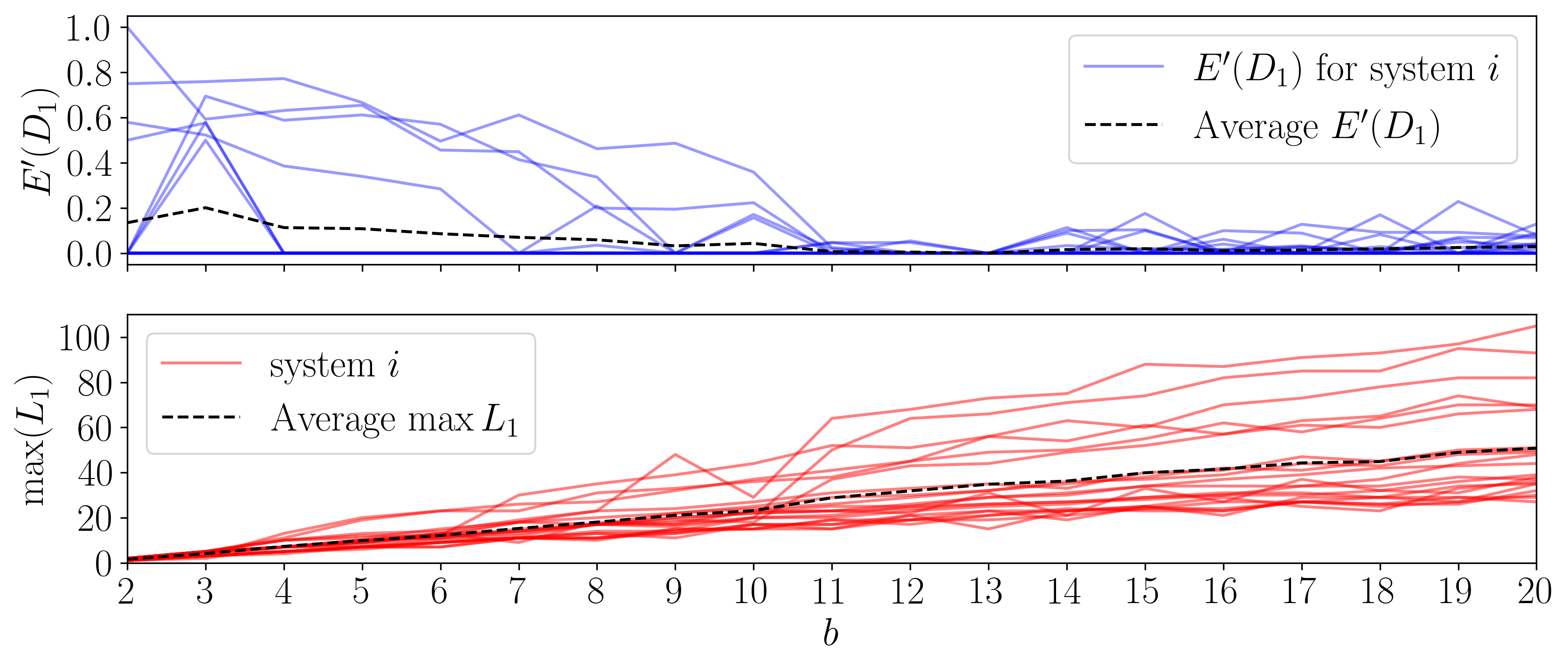}
	\caption{Binning size analysis using the normalized persistent entropy $E'(D_1)$ and maximum lifetime $\max(L_1)$ for 23 dynamical systems listed in Table~\ref{tab:systems} with $b \in [2,20]$.
	}
	\label{fig:binning_analysis_all_systems}
\end{figure}
The next question we want to ask is if $b=12$ is a good option for other dynamical systems. To test this we again calculate the $E'(D_1)$ and $\max(L_1)$ for $b \in [2,20]$ for the 23 dynamical systems listed in Table~\ref{tab:systems}. Figure~\ref{fig:binning_analysis_all_systems} shows these statistics for all of the dynamical systems and it demonstrates that a choice of $b \in [11,13]$ does work well for all of the dynamical systems with a drop in $E'(D_1)$. Based on this seemingly universal choice of $b$ in this work we use $b=12$ unless otherwise stated.

\section{Data} \label{app:data}

In this work we heavily rely on a 23 dynamical systems commonly used in dynamical systems analysis. All of these systems are continuous flow opposed to maps. The 23 systems are listed in Table~\ref{tab:systems}. The equations of motion for each systems can be found in the python topological signal processing package \texttt{Teaspoon} under the module \textit{MakeData} \url{https://lizliz.github.io/teaspoon/}. Specifically, these systems are described in the dynamical systems function of the make data module~\cite{Myers2020}.
\begin{table}
\centering
\caption{Continuous dynamical systems used in this work.}
\label{tab:systems}
\begin{tabular}{ll}
\textbf{Autonomous Flows} & \textbf{Driven Dissiptive Flows} \\ \hline
Lorenz & Driven Van der Pol Oscillator \\
Rossler & Shaw Van der Pol Oscillator \\
Double Pendulum & Forced Brusselator \\
Diffusionless Lorenz Attractor & Ueda Oscillator \\
Complex Butterfly & Duffing Van der Pol Oscillator \\
Chen's System & Base Excited Magnetic Pendulum \\
ACT Attractor &  \\
Rabinovich Frabrikant Attractor &  \\
Linear Feedback Rigid Body Motion System &  \\
Moore Spiegel Oscillator &  \\
Thomas Cyclically Symmetric Attractor &  \\
Halvorsen's Cyclically Symmetric Attractor &  \\
Burke Shaw Attractor &  \\
Rucklidge Attractor &  \\
WINDMI &  \\
Simplest Cubic Chaotic Flow & \\
\hline
\end{tabular}
\end{table}

Each system was solved to have a time delay $\tau = 50$, which was estimated from the multiscale permutation entropy method~\cite{Myers2020a}. The signals were simulated for $750 \tau /f_s$ seconds with only the last fifth of the signal used to avoid transients. It should be noted that we did not need to normalize the amplitude of the signal since the ordinal partition network is not dependent on the signal amplitude.

\section{Additional Results} \label{app:SVM_shortest_path_distances}
Here we provide the additional SVM projections to visualize the dynamic state detection performance of the shortest path distances: unweighted shortest path, shortest weighted path, and weighted shortest path. Table~\ref{tab:accuracies} provides the corresponding average accuracies.

\begin{figure}[ht]
    \centering
    
    \begin{minipage}[t]{0.32\textwidth}
        \centering
        \includegraphics[width=\linewidth]{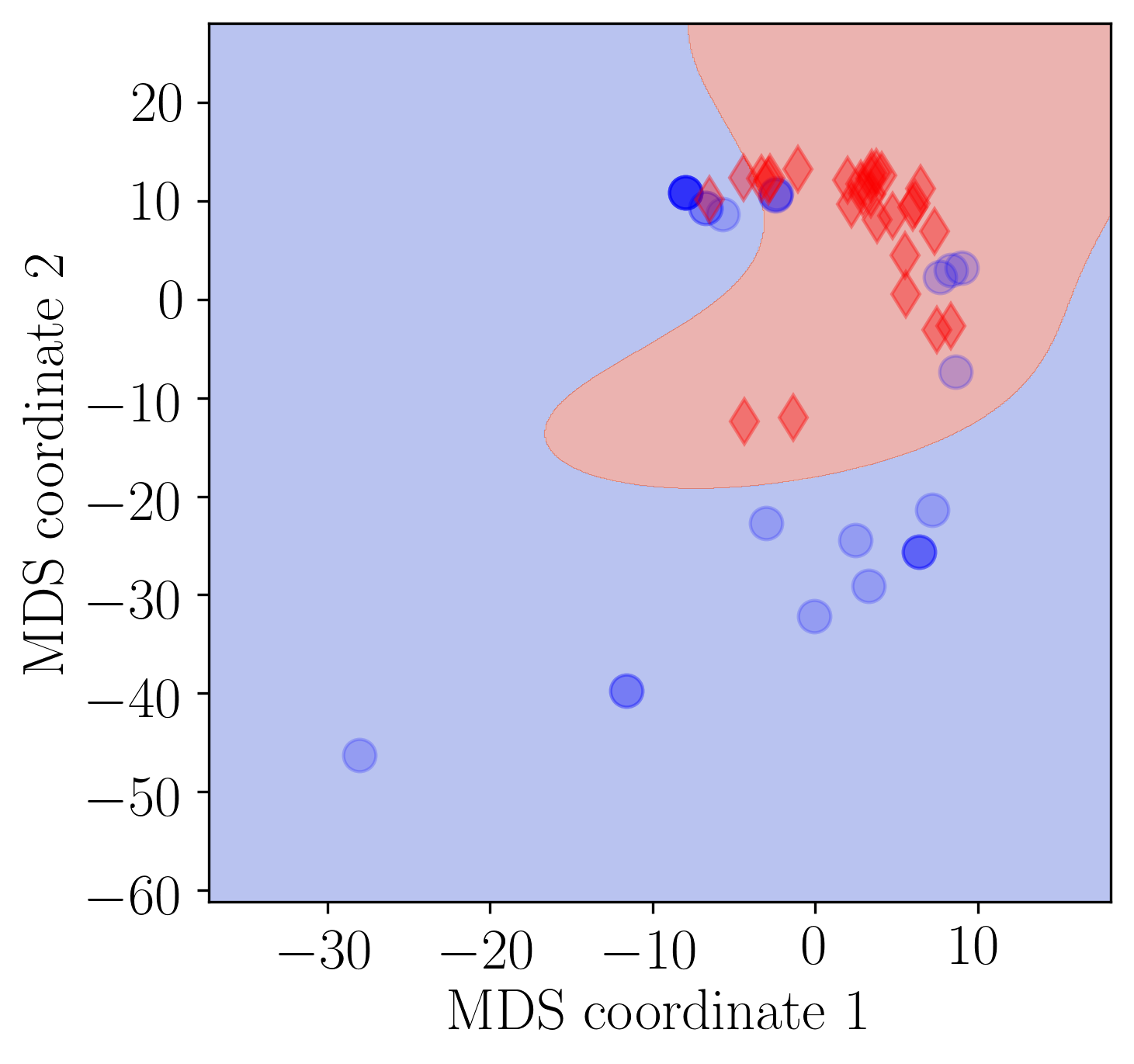} 
         {(a)}
    \end{minipage}
    \hfill
    \begin{minipage}[t]{0.32\textwidth}
        \centering
        \includegraphics[width=\linewidth]{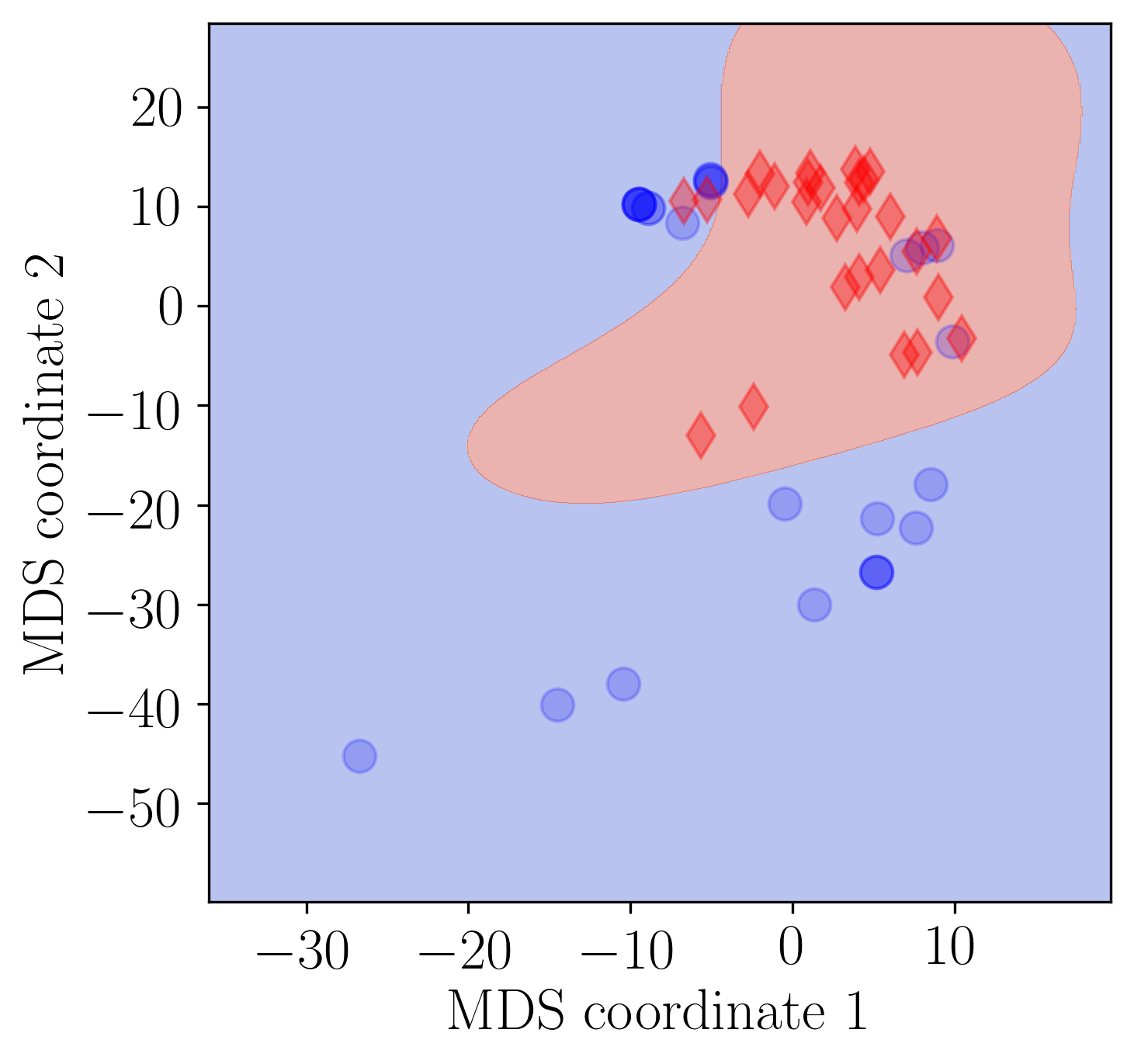} 
         {(b)}
    \end{minipage}
    \hfill
    \begin{minipage}[t]{0.32\textwidth}
        \centering
        \includegraphics[width=\linewidth]{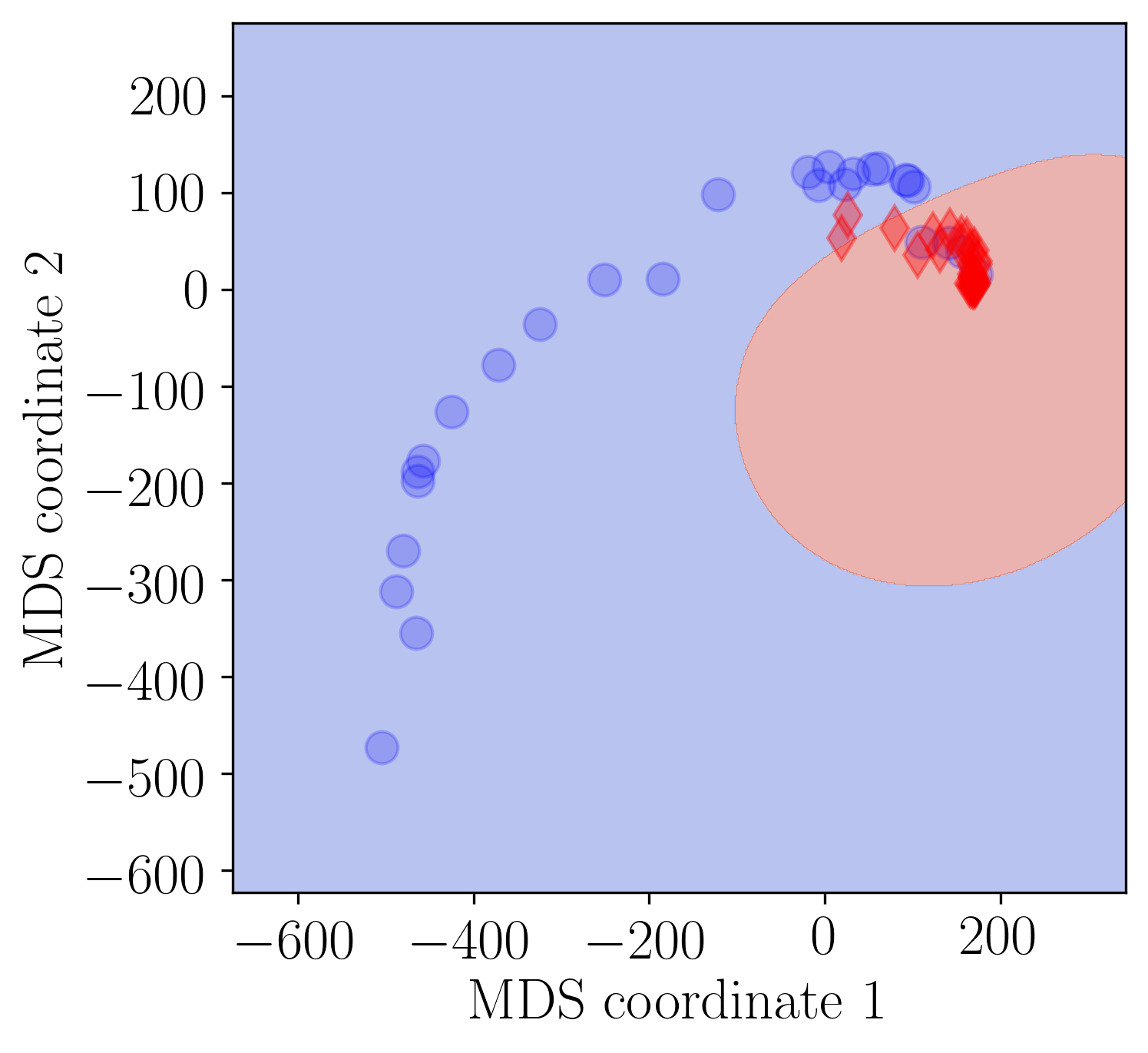} 
         {(c)}
    \end{minipage}
    \hspace{10mm}
    
    \begin{minipage}[t]{0.32\textwidth}
        \centering
        \includegraphics[width=\linewidth]{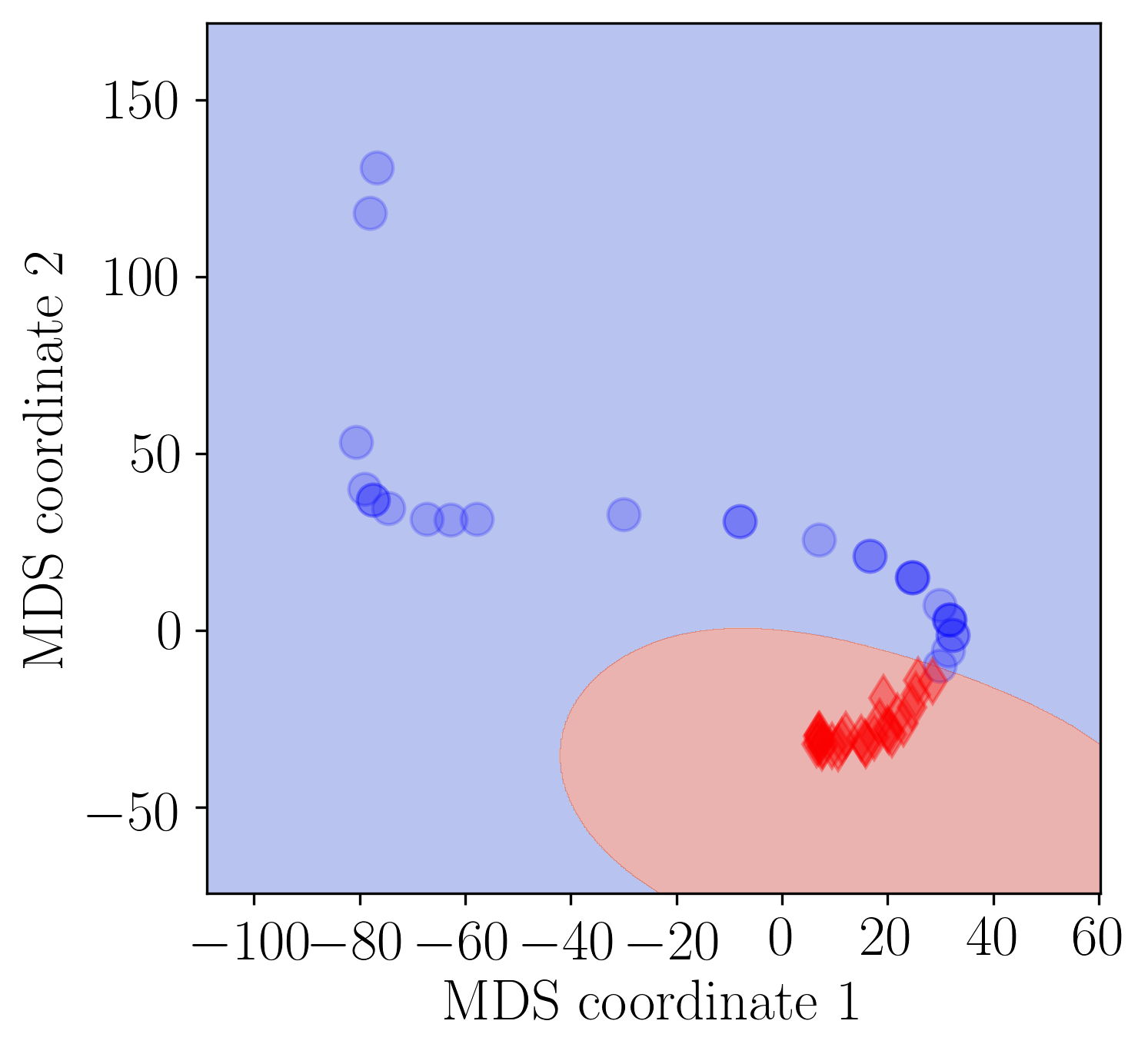} 
         {(d)}
    \end{minipage}
    \hfill
    \begin{minipage}[t]{0.32\textwidth}
        \centering
        \includegraphics[width=\linewidth]{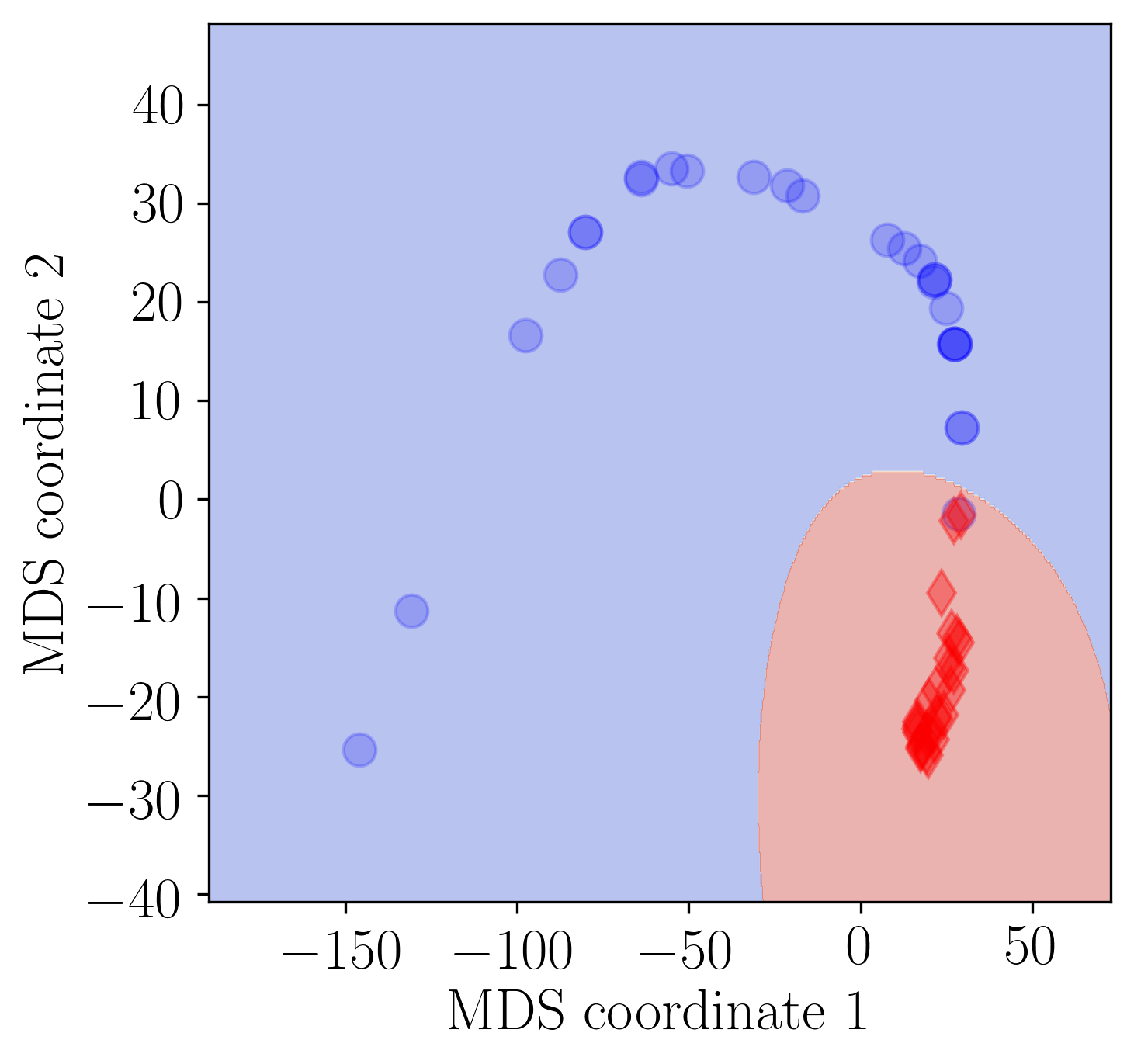} 
         {(e)}
    \end{minipage}
    \hfill
    \begin{minipage}[t]{0.32\textwidth}
        \centering
        \includegraphics[width=\linewidth]{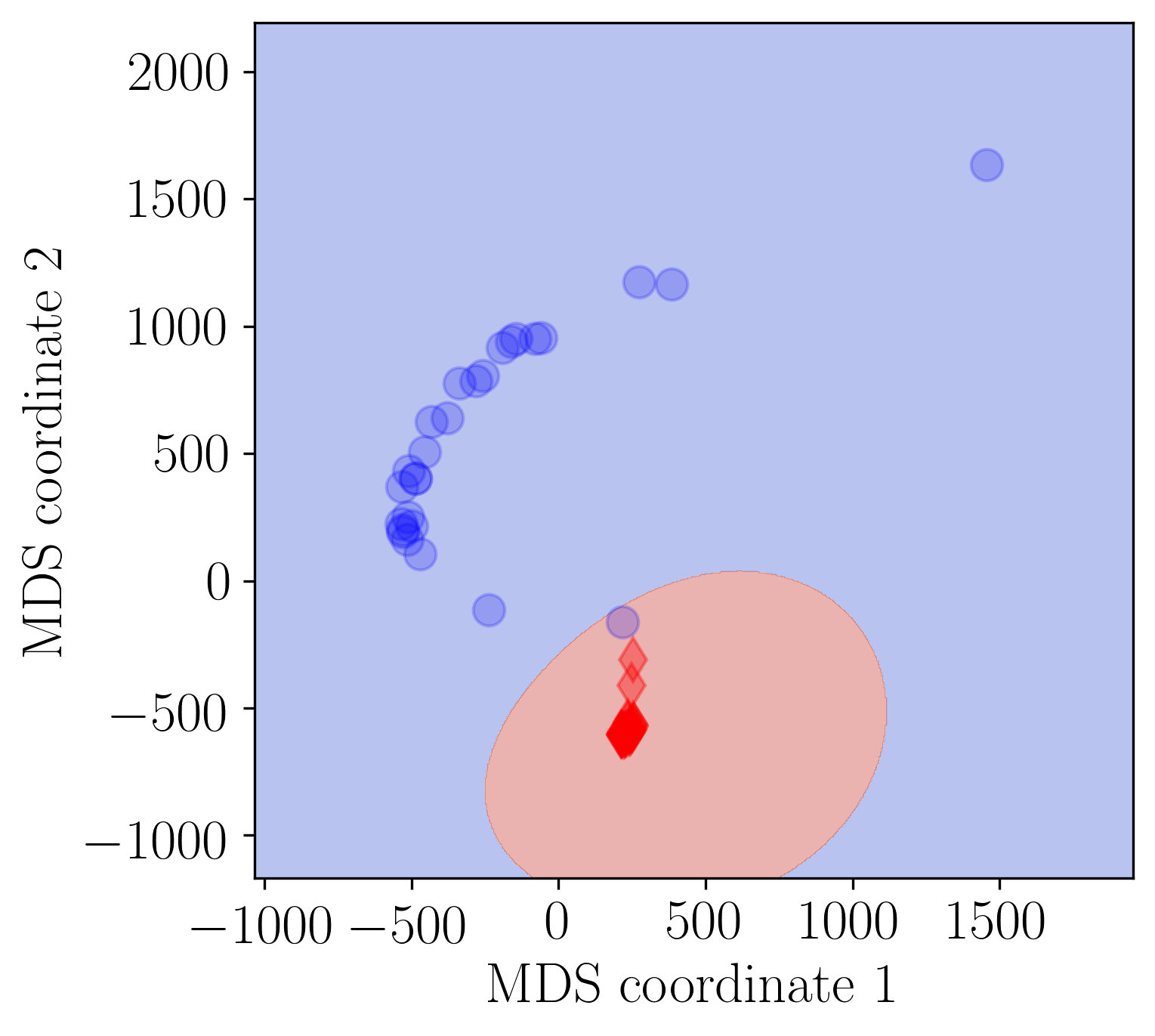} 
         {(f)}
    \end{minipage}
    
    \caption{Two dimensional MDS projection of the bottleneck distances between persistence diagrams of the chaotic and periodic dynamics with an SVM radial bias function kernel separation. This separation analysis was repeated for the OPNs and CGSSNs using the unweighted shortest path, shortest weighted path, and weighted shortest path distances. (a) Unweighted shortest path distance of OPN, (b) Shortest weighted path distance of OPN, (c) Weighted shortest path distance of OPN, (d) Unweighted shortest path distance of CGSSN, (e) Shortest weighted path distance of CGSSN, (f) Weighted shortest path distance of CGSSN.}
    \label{fig:MDS}
\end{figure}

\end{document}